\def\confName{ICCV}
\def\confYear{2023}

\def\paperTitle{STEPs: Self-Supervised Key Step Extraction and Localization from Unlabeled Procedural Videos}

\def\authorBlock{
    Anshul Shah$^1$ \qquad
    Benjamin Lundell$^2$ \qquad
    Harpreet Sawhney$^2$ \qquad 
    Rama Chellappa$^1$  \\ 
    \\
    $^1$ Johns Hopkins University \qquad $^2$ Microsoft Mixed Reality \\
    {\tt\small \{ashah95, rchella4\}@jhu.edu \{benjamin.lundell,harpreet.sawhney\}@microsoft.com}
}

\newif\ifreview 
\newif\ifarxiv \newcommand{\arxiv}{\arxivtrue}
\newif\ifcamera 
\newif\ifrebuttal 

\arxiv

\pdfoutput=1
\documentclass[10pt,twocolumn,letterpaper]{article}
\ifreview \usepackage[review]{cvpr} \fi
\ifarxiv \usepackage[pagenumbers]{cvpr} \fi
\ifrebuttal \usepackage[rebuttal]{cvpr} \fi
\ifcamera \usepackage{cvpr} \fi

\usepackage{graphicx}
\usepackage{amsmath}
\usepackage{amssymb}
\usepackage{booktabs}

\usepackage{times}
\usepackage{microtype}
\usepackage{epsfig}
\usepackage[table,xcdraw]{xcolor}
\usepackage{caption}
\usepackage{float}
\usepackage{placeins}
\usepackage{color, colortbl}
\usepackage{stfloats}
\usepackage{enumitem}
\usepackage{tabularx}
\usepackage{xstring}
\usepackage{multirow}
\usepackage{xspace}
\usepackage{url}
\usepackage{subcaption}
\usepackage{xcolor}
\usepackage[hang,flushmargin]{footmisc}
\usepackage{algorithm}
\usepackage{algorithmic}
\usepackage{enumitem,amssymb}
\usepackage{comment}
\usepackage{listings}
\usepackage{etoolbox}
\makeatletter
\AfterEndEnvironment{algorithm}{\let\@algcomment\relax}
\AtEndEnvironment{algorithm}{\kern2pt\hrule\relax\vskip3pt\@algcomment}
\let\@algcomment\relax
\newcommand\algcomment[1]{\def\@algcomment{\footnotesize#1}}
\renewcommand\fs@ruled{\def\@fs@cfont{\bfseries}\let\@fs@capt\floatc@ruled
  \def\@fs@pre{\hrule height.8pt depth0pt \kern2pt}%
  \def\@fs@post{}%
  \def\@fs@mid{\kern2pt\hrule\kern2pt}%
  \let\@fs@iftopcapt\iftrue}
\makeatother

\ifcamera \usepackage[accsupp]{axessibility} \fi

\ifarxiv  \fi

\makeatletter
\newcommand*\bigcdot{\mathpalette\bigcdot@{.5}}
\newcommand*\bigcdot@[2]{\mathbin{\vcenter{\hbox{\scalebox{#2}{$\m@th#1\bullet$}}}}}
\makeatother

\newcommand{\ours}{STEPs}
\newcommand{\Ours}{STEPs}

\newlist{todolist}{itemize}{2}
\setlist[todolist]{label=$\square$}
\usepackage{pifont}
\newcommand{\cmark}{\ding{51}}%
\newcommand{\xmark}{\ding{55}}%

\usepackage[normalem]{ulem}

\usepackage{xr-hyper}

\makeatletter
\newcommand*{\addFileDependency}[1]{
  \typeout{(#1)}
  \@addtofilelist{#1}
  \IfFileExists{#1}{}{\typeout{No file #1.}}
}

\makeatother

\usepackage[pagebackref,breaklinks,colorlinks]{hyperref}
\usepackage[capitalize]{cleveref}
\crefname{section}{Sec.}{Secs.}
\crefname{table}{Table}{Tables}
\crefname{figure}{Fig.}{Figs.}

\begin{document}
\title{\paperTitle}
\author{\authorBlock}
\maketitle

\begin{abstract}
We address the problem of extracting key steps from unlabeled procedural videos, motivated by the potential of Augmented Reality (AR) headsets to revolutionize job training and performance. 
We decompose the problem into two steps: representation learning and key steps extraction. We propose a training objective, Bootstrapped Multi-Cue Contrastive (BMC2) loss to learn discriminative representations for various steps without any labels. Different from prior works, we develop techniques to train a light-weight temporal module which uses off-the-shelf features for self supervision. Our approach can seamlessly leverage information from multiple cues like optical flow, depth or gaze to learn discriminative features for key-steps, making it amenable for AR applications. 
We finally extract key steps via a tunable algorithm that clusters the representations and samples.  
We show significant improvements over prior works for the task of key step localization and phase classification. Qualitative results demonstrate that the extracted key steps are meaningful and succinctly represent various steps of the procedural tasks. Our code can be found at \href{https://github.com/anshulbshah/STEPs}{https://github.com/anshulbshah/STEPs}.
\end{abstract}
\section{Introduction}
\label{sec:intro}
Rapid shifts in technology and business models have led to a mismatch between the skills needed by employers and the skills possessed by the labor force. It has been estimated that this mismatch will reduce manufacturing output by \$2.4 trillion over ten years in the US alone \cite{giffi2018deloitte, guldimann2022green}. Increased attention has been placed on effective methods of ``reskilling'' workers \cite{agrawal2020beyond}. Unfortunately, reskilling will not be easy: human expertise in performing a complex task takes years of training and mastery of domain-specific knowledge \cite{bryan1899studies, ericsson1996expert}.

Augmented Reality (AR) headsets can play an important role in collective reskilling efforts. AR headsets are known to improve the efficiency of front line workers during training and on the job, across industries as diverse as food service, manufacturing, medicine, and warehousing\cite{abraham2017augmented, clark2019educational, ruthberg2020mixed, schwerdtfeger2009pick}. AR plays a strikingly similar role across these diverse use cases: to assist the user in completing a complex task, the headset renders a sequence of visual cues on real-world objects.  

Our approach focuses on extracting key-steps of a complex task which is the most crucial component needed for automatic AR content creation.
We employ a ``learning-from-observation''-style framework \cite{michalski1983learning}, where an instructor is recorded while performing a complex task. The goal is to automatically parse the recording into {\em key steps} (KSs) that succinctly represent the complete task. This greatly streamlines the content creation process, as the trainer no longer has to manually edit the recording to find the key steps.

\begin{figure}
    \centering
    \includegraphics[trim={0 4cm 0 4cm},clip,width=\linewidth]{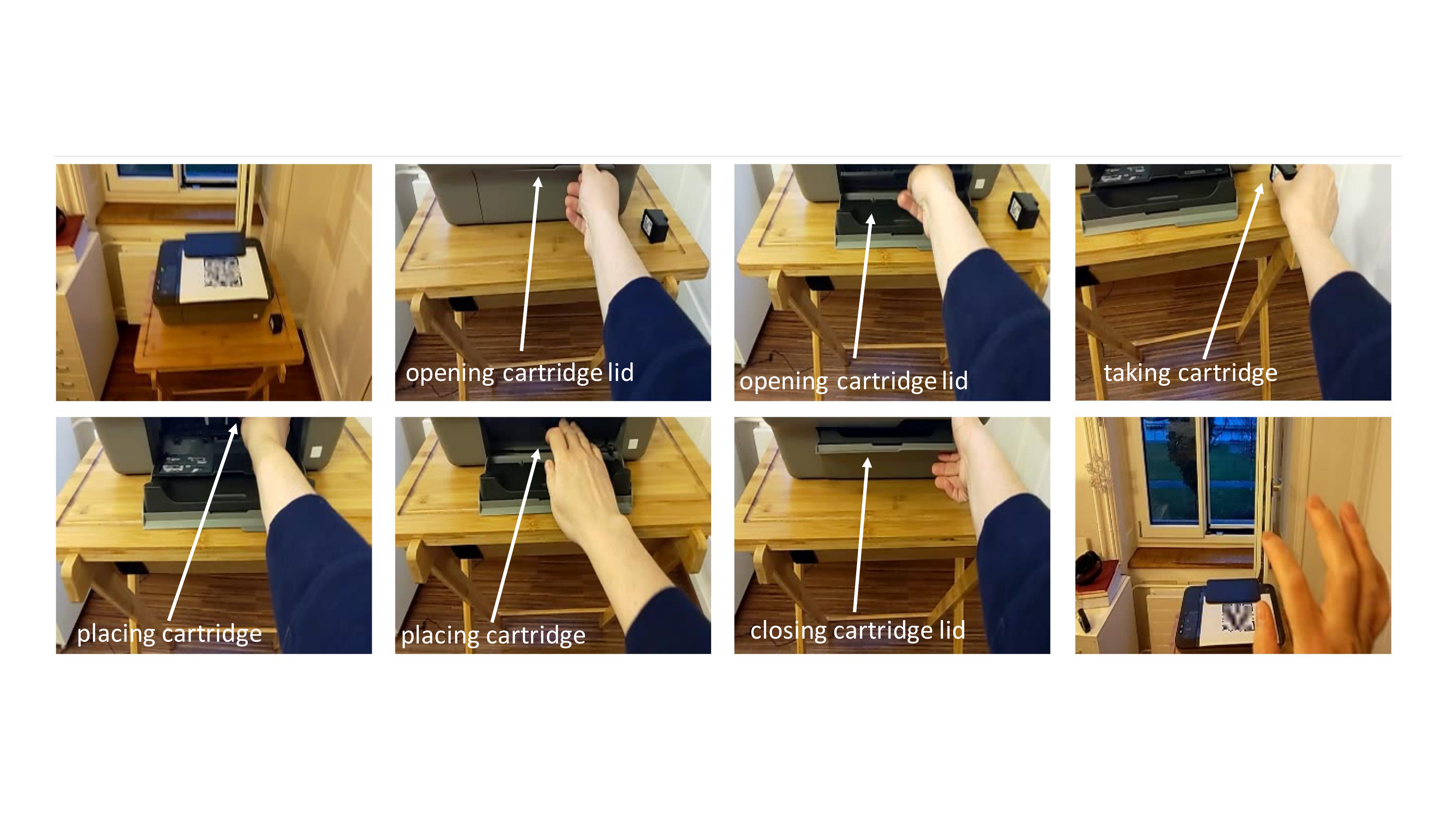}
    \caption{An illustrative example of automatically generated key steps using our approach \ours{} for the task of changing the cartridge in a printer. Data
was captured using a Microsoft HoloLens 2 and extracted using a publicly available repository. Our approach leads to plausible key steps for sub-tasks of `opening cartridge lid', `taking cartridge', `placing cartridge' and `closing cartridge lid` and `end recording'. Note that the associated labels and arrows are for visualization purposes only and were not used for training. }
\vspace{-0.5cm}
    \label{fig:teaser_figure}
\end{figure}

Consider the task of changing the cartridge in a printer. Using tools from the public repository \cite{hl2_rm}, we captured data on a Microsoft HoloLens 2 of an ``expert'' undertaking this task. Using multiple cues -- hand pose, head pose, eye gaze and first person video -- we automatically generate the key-steps shown in Figure ~\ref{fig:teaser_figure}. Using multiple cues when observing experts perform procedural tasks is important when generating training materials for novices~\cite{charness2008role,gopher1994transfer, fadde2018training}.

The key step extraction problem for complex procedural tasks is challenging: (1) Recordings of tasks performed by experts are limited in number; (2)  Supervision for key steps is hard due to the subjective nature of what constitutes a key step; (3) There are no large-scale datasets for real world procedural tasks. Unlike typical web-crawled videos used in video representation learning, procedures are often minutes long; and (4) Visual information alone might not adequately represent all the information in a scene. None of the prior works address all of the challenges mentioned above. 

Our solution involves a two-stage approach to the KS extraction problem. First, we train a task-specific model to produce a context-rich feature vector for each frame of the recording without any labels. To overcome paucity of data, our Multi-cue key steps (\ours{}) approach, (Figure \ref{fig:feature_extraction} and \cref{sec:method}), uses multiple temporal feature sequences corresponding to different cues as input. Different from prior works in self-supervised learning, we only train a per-modality temporal model which is applied on pre-extracted `raw' features using a pre-trained feature extractor. 
To learn rich features, we propose a novel loss termed BMC2 (Bootstrapped Multi-Cue Contrastive Loss). The loss enforces temporally adapted features from each modality within the window to be close in a common latent space. Instead of using a fixed temporal window as used by related works, our loss first bootstraps a temporal window around each anchor using `raw' features thus reducing the effect of false negatives. After training, the per-frame representations are clustered and KSs are sampled from the clusters. Use of off-the-shelf features enables fast temporal encoder training with extremely long temporal contexts; we can train a model on 17 Meccano bike assembly videos~\cite{ben2021ikea} for 300 epochs in under 6 minutes on a single GPU. 
Unlike recent works in self-supervised video representation learning \cite{dwibedi2019temporal,haresh2021learning, liu2021learning}, we do \emph{not} rely on alignment between multiple recordings as a proxy task. {\em Inter-video alignment} as a learning loss requires multiple recordings of the same task, thus limiting its practical applicability to the problem  of learning from few expert recordings.

Our approach is suitable for not only multi-modal data computed by first-person wearable devices such as Oculus Quest, Microsoft HoloLens and others, but also conventional procedural task videos.
In this case, we use the visual stream to extract rich targeted modalities like optical flow for training. 

We provide quantitative and qualitative results on seven datasets (both first-person and third person). Unfortunately, KS extraction is a subjective task and there are no publicly available large-scale annotated procedural learning datasets to evaluate KSE. Thus for quantatitive benchmarking, we compare our learned representations for the proxy task of Key Step localization (KSL). On that task, we outperform \cite{bansal2022my} by 3-18 points on F1 score.  
We also provide results on phase classification and outperform prior works by a wide margin. 

In summary, following are our key contributions:
\begin{enumerate}
  \setlength{\itemsep}{0pt}
  \setlength{\parskip}{0pt}
    \item We propose an intra-video SSL-approach for procedure learning to train a light-weight temporal encoder with pre-extracted multi-cue features.  We can train on long temporal sequences (1000s of frames) unlike prior works that train expensively large backbones.
    \item We employ a novel loss, termed BMC2, that bootstraps information from pre-extracted multi-cue features to create a temporal window around an anchor, which is used to force representations from multiple modalities to be close in a common representation space.
    To the best of our knowledge, ours is the first work to employ multiple synchronized sensor and vision-derived cues often available in AR devices for key step extraction. 
    \item We obtain superior performance compared to state-of-the-art on several tasks across various datasets: KS localization, phase classification and Kendall's Tau with a low-complexity SSL module. 
\end{enumerate}

\section{Related Work}
\label{sec:related}

\ours{} is inspired by current video understanding research including KS localization, procedure learning, self-supervised learning and video summarization.

\textbf{KS extraction \& procedure learning of complex tasks.}
To obtain key steps, prior works use a state transition model \cite{elhamifar2019unsupervised}, Mallows model ~\cite{sener2018unsupervised}, clustering and ordering of visual features ~\cite{kukleva2019unsupervised} or weak alignment between visual and linguistic cues~\cite{shen2021learning}. These approaches either rely on per-task training or additional language cues. A subset selection module is used as a teacher in \cite{elhamifar2020self} to obtain an unsupervised localization of KSs in a multi-task setting assuming access to task labels. Approaches that aim to learn features from complex task videos use pretext tasks to learn representations without supervision. 
They learn features using pairs of videos from a task by finding correspondences and imposing constraints like cycle-consistency~\cite{dwibedi2019temporal}, time-warping~\cite{haresh2021learning} and optimal-transport based alignment~\cite{liu2021learning}. Recently, ~\cite{bansal2022my} proposed to use a loss function that builds over advances in video alignment and temporal consistency to learn strong models for procedure learning on both ego-centric and third person procedural datasets. 
Different from these prior works, we adapt off-the-shelf features and do not require pairs of videos in training. Multiple cues in training provide significant improvement over the state-of-the-art. We also propose an approach to improve the set of positives used to train models with temporal consistency. Our approach also works in a low-shot setting since it does not rely on alignment. 

\noindent \textbf{SSL for video understanding.} Contrastive learning (CL) ~\cite{gutmann2010noise,hadsell2006dimensionality,oord2018representation} shows impressive improvements on image-based self-supervision \cite{chen2020simple,he2020momentum,tian2020contrastive,li2020prototypical,robinson2020hard,shah2022max}. Videos allow for additional constraints for self-supervision like discriminating temporally transformed version of the video~\cite{jenni2020video}, predicting speediness \cite{epstein2020oops,benaim2020speednet,yao2020video} or use of multiple camera views~\cite{shah2023multi,das2023viewclr}. Some works use alternate pretexts~\cite{kim2019self} while others employ temporal coherence and ordering as signals \cite{misra2016shuffle,buchler2018improving,wei2018learning,lee2017unsupervised,xu2019self,hu2021contrast}. 
Most video-based SSL approaches focus on learning from short, trimmed videos depicting a single action. Real world task videos are rarely short or trimmed. To address this, \cite{zhukov2020learning} uses order verification to isolate actions from background, using global context and segment-based regularization~\cite{kuang2021video}, exploiting relations between clips~\cite{luo2022exploring} or devising an action boundary sensitive pretext task~\cite{xu2021boundary}.
Unlike typical SSL models trained on a large collection of videos, we work with training on a few videos. Instead of training a feature extractor, we use off-the-shelf features and train a low complexity temporal model. 

\noindent \textbf{Multiple modalities for SSL:} Multi-modal (e.g. audio and text) methods use cross-modal losses and augmentations \cite{miech2019howto100m,patrick2021compositions,morgado2021audio,recasens2021broaden,zolfaghari2021crossclr,wang2021fine,pramanick2022volta} to learn features. Use of additional modalities/cues derived directly from the visual stream have been very successful for supervised learning~\cite{coskun2021domain,shah2022pose,duan2022revisiting}. Optical flow has been used by various works \cite{han2020self,xiao2021modist,xiong2021multiview,li2021motion} to learn and distill information from the motion stream while \cite{rai2021cocon} explores cooperative CL for multi-modal SSL but for training a feature encoder on short trimmed videos. 
SSL relies on data augmentations which are not trivial when working with pre-extracted features. Working with multiple cues helps solve this problem by using contrastive learning similar to the ones used in approaches discussed above but without any explicit data augmentations. MM-SADA~\cite{munro2020multi} proposed a modality alignment loss assuming access to trimmed action segments and applied it for domain adaptation. Our loss function is inspired by CoCLR\cite{han2020self} but differs in that  our loss objectives employ temporal sequence of features instead of one feature per video. Further, we work with a CIDM-based objective and propose improvements to it. Different from these prior works, we do not aim to learn a feature extractor but a temporal model. 
We also note that natural language modality has been extensively used in procedure learning and related tasks ~\cite{alayrac2016unsupervised,doughty2020action,fried2020learning,malmaud2015s,sener2018unsupervised,shen2021learning,yu2014instructional,lin2021exploring,lin2022egocentric, pramanick2023egovlpv2,dvornik2023stepformer,ashutosh2023video}. Instead, our work explores the use of other modalities: on-device sensors like gaze and depth which are often available on AR devices or vision-derived modalities like motion or human pose. Language is only weakly aligned with text for procedural videos and might not be always available.

\textbf{Unsupervised video summarization}
Video summarization without labels is another relevant thread for our work. Past works have used Reinforcement learning \cite{yoon2021interp,zhou2018deep}, subset selection~\cite{shemer2021ils}, clustering \cite{shroff2010video,turaga2009unsupervised,dhamecha2020video,jadon2020unsupervised} and GANs \cite{mahasseni2017unsupervised,yuan2019unsupervised,apostolidis2020ac,apostolidis2020unsupervised,he2019unsupervised} or SSL~\cite{gao2020unsupervised} to extract summaries from videos. 
In contrast, KSs are task based and require understanding the video at an implicit semantic level. Our target use case does not have videos with abrupt changes, shot changes or extreme camera motion all of which can be cues for determining key steps. Instead we deal with videos which often have subtle changes as a task is performed. Our approach can be used for other high-level vision tasks such as phase classification. That said, our learned features can be used with existing unsupervised video summarization approaches and alternative clustering based approaches.

\section{Approach}
\label{sec:method}
A straightforward way of extracting key-steps would have been to use supervised learning to classify each frame of a video to KS or not a KS. Unfortunately, there do not exist any large scale procedural datasets with annotated key steps. One could use self-supervised learning to extract rich features from these datasets followed by finetuning on a small carefully labeled set. But since these videos are recorded by experts, even having a large unlabeled dataset is impractical. This is a problem since self-supervised learning has been shown to be data hungry. Thus we propose to use a two-stage solution of self-supervised representation learning followed by a cluster-and-sample pipeline for key step extraction. This entire pipeline does not use \emph{any} labels to determine key steps. Next, we describe our approach assuming $M$ modalities ($m_i, i \in [1, \cdots M]$).
 
\subsection{Extracting raw features} \label{sec:extractraw}
The first step involves extracting raw features for use in the training pipeline. Our feature extractor could be any off-the-shelf backbone network like ResNet50~\cite{he2016deep}, RAFT encoder~\cite{teed2020raft}, human pose extractor or even on-device modalities like head pose, hand articulations, gaze and object pose~\cite{hl2_website}. The idea of adapting pre-extracted off-the-shelf features enables us to effectively leverage SSL on extremely small datasets. 

During training, we divide the entire video into $N$ chunks and randomly sample a frame from each chunk. Note that since our approach works with features instead of videos/images, we cannot employ any of the input-space augmentations. We apply temporal augmentations on features but do not generate multiple augmentations of the raw video. 

\begin{figure}
    \centering
    \includegraphics[width=\linewidth]{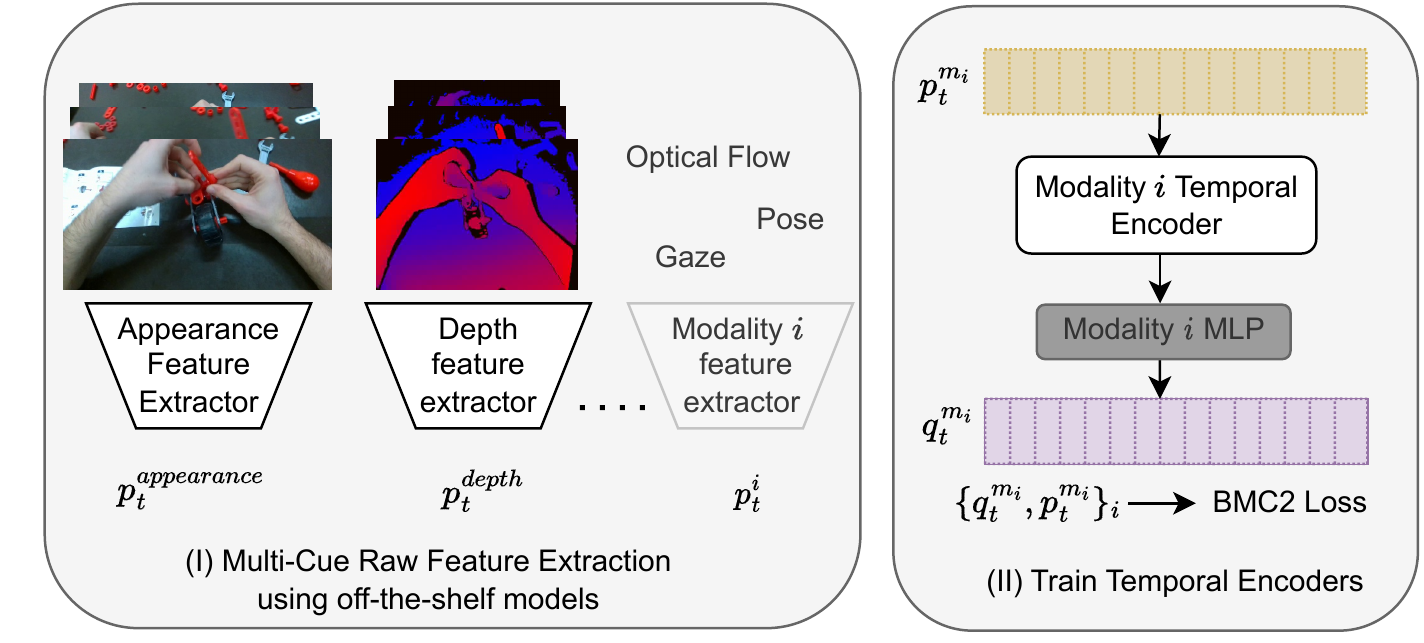}
    \caption{\ours{} pipeline. (I) We first extract framewise multi-cue features ($p^{m_i}_t$) for an input video using off-the-shelf multi-cue feature extractors. (II) these features are then adapted to the task using a temporal feature encoder which returns framewise adapted features ($q_t$). We propose to use our BMC2 loss which enforces continuity in time and cross-cue contrastiveness to learn rich features. These features are then used with a variety of tasks including cluster + sampling for key-step extraction, key-step localization or phase classification.}
    \label{fig:feature_extraction}
\end{figure}

The resultant features are denoted as $p_{t}^{m_i} \in \mathbb{R}^{D_i}$. We call these raw per-frame features (\cref{fig:feature_extraction}).

\subsection{Encoding temporal context} \label{sec:extracttemporal}
The objective of our model is to take the raw per-frame features and adapt them so that these are useful for various downstream tasks. We use a temporal encoder (vanilla transformer) to capture the long-range, temporal dynamics of an instructional video. Specifically, the raw sequences $p_{t}^{m_i}$ are input to a per-modality temporal encoder to generate a sequence of adapted per-frame features $\tilde{q}_{t}^{m_i}$. Finally, a per-modality MLP extracts a projection of the adapted sequence and L2 normalizes it to obtain ${q}_{t}^{m_i} \in \mathbb{R}^D$.

\subsection{Bootstrapped Multi-Cue Contrastive Loss}
\begin{figure}
    \centering
    \includegraphics[width=0.9\linewidth]{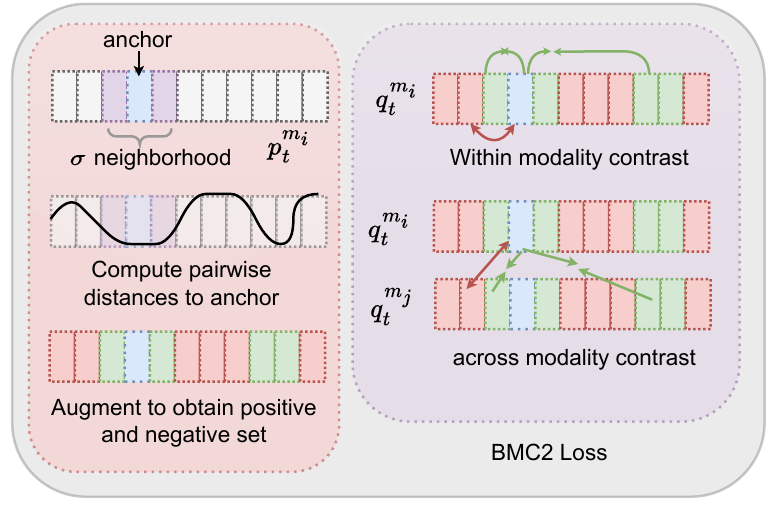}
    \caption{BMC2 loss. (Left) We start with a $\sigma$-neighborhood of the anchor and compute pairwise distances to other frame features of the video. We compute the mean distance within the neighborhood and use this as a threshold to find potential false negatives outside the window to obtain a new augmented positive set. (Right) We apply a multicue contrastive loss which enforces that positives are close in latent space across modalities.}
    \label{fig:loss_figure}
\end{figure}

Ideally, we want to learn representations such that features corresponding to all frames of a step (like attaching the leg of a table) lie close in the latent space. 
Since we do not have access to labels, we instead rely on a temporal consistency-based pretext task: features for frames close in time (positives) must be close in latent space while being far away from features of temporally far steps. Our contrastive learning objective is based on CIDM loss~\cite{haresh2021learning}. While this has been explored in the past~\cite{bansal2022my, liu2021learning,haresh2021learning}, in this work we propose two improvements. 
We first define the vanilla loss and then describe our novel contributions to this loss. 

The CIDM loss is used to enforce temporal consistency in the features. For an anchor $a$, the loss is defined as follows: 
{\small
\begin{align}
    \label{eq:cidm}
    &\mathcal{L}_{CL}(q_{t_a},q_{t_j},\mathcal{W}) = \mathcal{W}(t_a, t_j) \frac{d(q_{t_a}, q_{t_j})}{\gamma(t_a, t_j)} \notag \\ 
    &+ \left( 1 - \mathcal{W}(t_a, t_j) \right) \gamma(t_a, t_j) \max \left( 0, \zeta - d(q_{t_a}, q_{t_j}) \right)
\end{align}}
Here, \mbox{$\gamma(t_a,t_j) = (t_a - t_j)^2 + 1$}, $d(q_{t_a}, q_{t_j})$ is the Euclidean distance between $q_{t_a}$, $q_{t_j}$, $\zeta$ is the margin parameter, and $\mathcal{W}$ is the window such that, \mbox{$\mathcal{W}(t_i, t_j) = 1$} if $|t_i - t_j| \leq \sigma$, and $0$ otherwise. Here, $\sigma$ is the window size used to define the set of positives.

\begin{figure*}
    \centering
    \includegraphics[width=\linewidth]{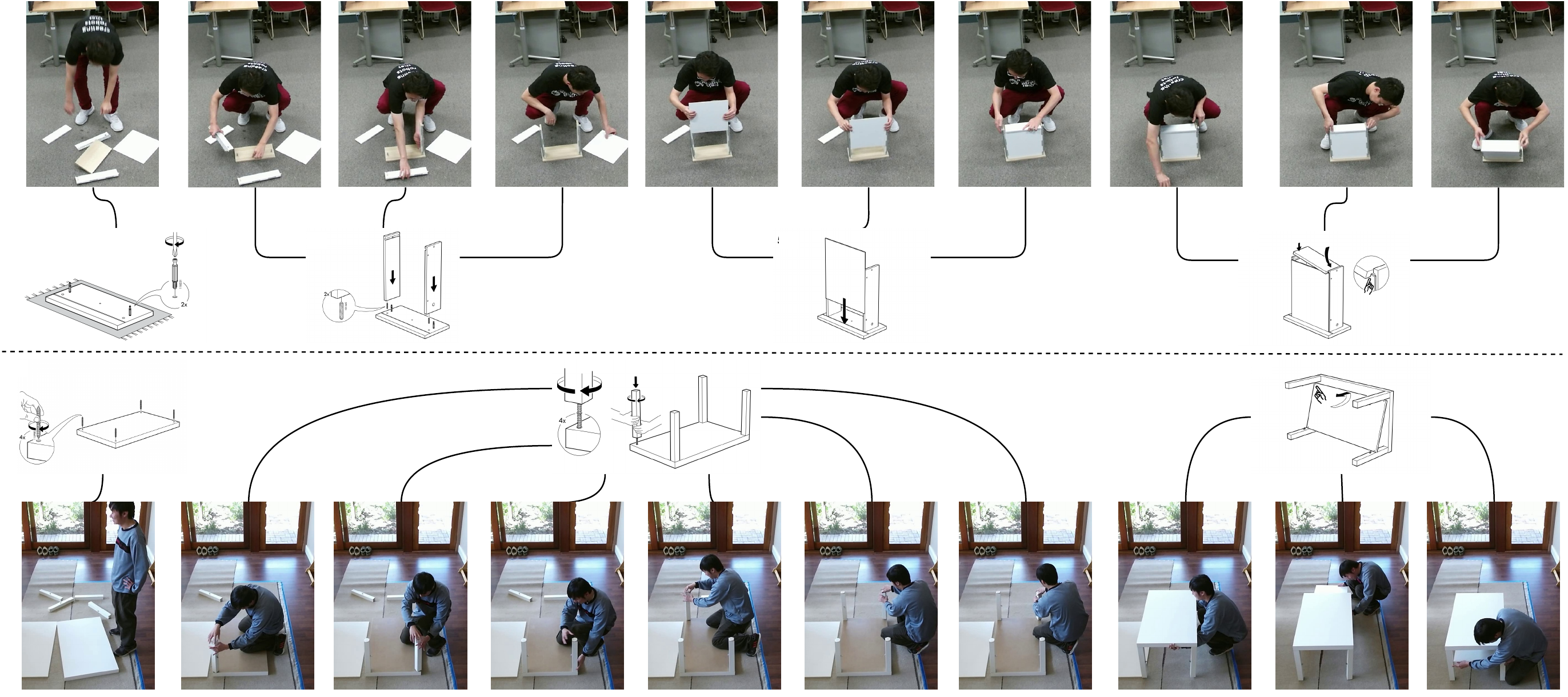}
    \caption{Key steps generated using \ours{}. We show results on two videos from the assembly of `Kallax Shelf Drawer' (top) and `Lack Coffee Table' (bottom) respectively. We see that they generated key steps are plausible given the complex task. We also manually match the generated steps to the drawings in IKEA user manual and we see that that the steps correlate well with the manual. Note that some of our generated steps get matched to a single step in the manual since steps in the manual are often over simplifications.}
    \label{fig:km_qualitative}
\end{figure*}
While this loss has been successfully applied in prior works, it has certain issues. Consider the task of following a recipe. Chopping certain vegetables might repeat across the video. Using a fixed temporal window, could use two time-separated instances of the same step as negatives instead of positives thus giving an incorrect training signal. We propose an approach to fix this issue and expand the temporal window. 

\noindent\textbf{Temporal window generating function:} The key idea is to find potential false negatives for a given anchor $a$ by bootstrapping on the raw features. Intuitively, the raw features in different modalities should have enough signal to discard some false negatives and use these as positives instead thus improving the training signal. Given an anchor $a$, we have its  raw features as $p_{t_a}^{m_i}$ and those of the other time steps as $p_{t_j}^{m_i} j \neq a$.  We first calculate the embedding distance of each frame from the anchor. Next, we use the mean distance within the $\sigma$ window as a threshold $\delta$ and recover time instances which have feature distances within that threshold. Thus, we have the following new window function: 
{
\small
\begin{align}
        \mathcal{W'}(t_a, t_j)= 
\begin{cases}
    1,& \text{if } d(p_{t_a}^{m_i},p_{t_j}^{m_i}) \leq \delta\\
    0,              & \text{otherwise}
\end{cases}
\end{align}}

\noindent We augment the original window to obtain $\tilde{\mathcal{W}} = \min(\mathcal{W'}$ + $\mathcal{W},1)$. 

\noindent\textbf{Multi-cue consistency loss:} In addition to enforcing temporal consistency between features from the same modality, we also enforce that features from different modalities within the window $\tilde{\mathcal{W}}$ are close in a common latent space. Our formulation easily allows us to do that by using $q_{t_a},q_{t_j}$ \cref{eq:cidm} from different modalities. 

\noindent\textbf{Overall BMC2 loss:}
We train our model with the weighted loss function : 

\[
    \mathcal{L}_{BMC2} = \sum_{u,v=1}^{M}\lambda_{uv}\mathcal{L}_{CL}(q^{m_u}_{t_a},q^{m_v}_{t_j},\tilde{\mathcal{W}})),
\]
where $\lambda_{uu} = 1$ for all $u$. \cref{fig:loss_figure} provides an illustration of the proposed loss function. 

Note that, we \emph{do not} back-propagate through the backbone feature extractors, $f^\star$, when training this pipeline. As the number of parameters in the temporal encoder and MLP modules is substantially smaller than the number of parameters in the backbone network (1.8M for our model vs  23M for ResNet-50), precomputing the features substantially reduces the size of the gradients during training, allowing us to process very long feature sequences on modest hardware. In turn, this enables us to extract rich temporal information from the videos while simultaneously making the training process efficient. In particular, the features $p_t$ are considered fixed throughout the training process and need only be computed once. They are completely determined by the choice of backbone network; we examine different choices of backbone networks in the supplementary material.

\subsection{Clustering and Sampling}
We follow a cluster-and-sample paradigm to sample KS once we have a trained model. Our approach borrows the clustering and ordering approach used in ~\cite{kukleva2019unsupervised} but adapts their segmentation approach to sample \emph{key steps} given a \emph{single} video. Since the learned representations are quite effective, even simple K-Means clustering algorithm is sufficient to obtain partitions of the aggregated feature data for a video, each partition potentially corresponding to a sub-task or a key step. Our approach allows for a choice of a clustering algorithm based on user constraints, including algorithms that do not require a fixed number of clusters to be specified in advance~\cite{finch}. To sample key steps, we first use a background frame rejection model~\cite{kukleva2019unsupervised} to reject steps which might correspond to background frames. Subsequently, each frame in the video is first assigned to one of the clusters. A step from each cluster is then chosen based on its distance to the cluster center and the key steps are temporally ordered. While sampling from a cluster, we perform an additional partitioning step based on the timestamp to ensure that we sample key steps of the same potential sub-task but happening at different time instances. Algorithm~\ref{alg:sampling_keymoments} details our approach to cluster and sample key steps.

\begin{algorithm}[!t]
\footnotesize
\caption{\label{alg:sampling_keymoments} Extracting Key steps}
\begin{algorithmic}
    \STATE \textbf{Input:} Per-frame adapted features for the candidate video $\{\tilde{q}_t\}, t\in[1,\cdots,T]$, num clusters K 
    \STATE \textbf{Output:} Key-steps $KS(v) =   \{a_k\}_{k=1}^{k=K}$, 
    \STATE \textcolor{gray}{\# Cluster the video}
    \STATE $\mathcal{C}=\texttt{cluster}(\{\tilde{q}_t\}, K)$ where $\mathcal{C}=\{C_i\}$
    \STATE \textcolor{gray}{\# Obtain assignment and distance to closest cluster}
    \STATE $\{y_t,d_t\}=\texttt{predict}\_\texttt{cluster}(\{\tilde{q}_t\},\mathcal{C})$
    \STATE \textcolor{gray}{\# Sampling key-steps}
    \STATE \textbf{for} cluster $k = 1, \ldots, K$ \textbf{do}
        \STATE $~~~~$ \textcolor{gray}{\# Obtain time indices for frames assigned to $\mathcal{C}_k$}
        \STATE $~~~~$ $\mathcal{T}_k = \{t ~| y_t = k~~ \forall t \in [1,\cdots,T] \}$
        \STATE $~~~~$ \textcolor{gray}{\# Remove elements farthest from the cluster center}
        \STATE $~~~~$ $\mathcal{T}'_k = \texttt{background}\_\texttt{reject}(\mathcal{T}_k,\alpha)$
        \STATE $~~~~$ \textcolor{gray}{\# Break into segments if adjacent indices are $\gamma$ away}
        \STATE $~~~~$ $\mathcal{T}''_k =  \texttt{split}\_\texttt{to}\_\texttt{segments}(\mathcal{T}'_k,\gamma)$
        \STATE $~~~~$ \textcolor{gray}{\# Sample key steps from each segment}
        \STATE $~~~~$ $a_k = \{t \!\sim\! \mathcal{T}''_{kj} \}$  
    \STATE \textbf{end for}
    \STATE return $\{a_k\}_{k=1}^{k=K}$
\end{algorithmic}
\end{algorithm}

\section{Experiments and Analyses}
\label{experiments}
In this section, we evaluate the efficacy of our representation learning pipeline. KS extraction is a subjective task and there do not exist any large scale datasets with labeled KS. Thus for quantatitive benchmarking, we compare our learned representations for the related task of Key Step localization (KSL). Furthermore, to demonstrate the usefulness of our SSL approach generally, we offer a quantitative comparison on other benchmarks used in the self-supervised video procedure learning for task based videos. Our qualitative visualizations of the succinct KS prove the efficacy of our approach. Finally, we provide analyses to understand the working of our approach. 

\noindent\textbf{Datasets:} 
We evaluate our approach on four egocentric procedure learning datasets, Meccano~\cite{ragusa2021meccano} (toy bike assembly), EPIC-Tents~\cite{jang2019epic} (camping tent assembly), CMU Kitchens~\cite{de2009guide} and EGTEA-Gaze+~\cite{li2018eye} (cooking recipes). Unlike prior work which use just the visual stream from these datasets, we also explore the use of modalities like depth and gaze available with these egocentric datasets during the training stage. Because they are captured from an egocentric viewpoint and have other synchronized sensor recordings, these datasets are very relevant to our motivating use case of AR-based instruction.
In addition to these, we also evaluate on three third-person procedural datasets, IkeaAssembly~\cite{ben2021ikea} (furniture assembly), CrossTask~\cite{zhukov2020learning} and ProceL~\cite{elhamifar2019unsupervised} which contain diverse procedures. We also evaluate on PennActions which while not a procedural dataset, is a common benchmark used in unsupervised video parsing works. 

\noindent\textbf{Implementation Details:} Our temporal encoder consists of a two layer vanilla transformer model. 
We use two modalities $M=2$ for all experiments in the main paper. 
Additional sensor modalities (like Gaze, Depth, etc) exist for some of these datasets but are not consistent across datasets. To tease out the effect of nature of sensor modalities from the learning approach, all comparisons with prior works use features extracted using ResNet-50 ($m_1$) and  RAFT-OpticalFlow~\cite{teed2020raft} ($m_2$) unless otherwise mentioned. We choose these two modalities since they can readily be extracted from any procedural task videos and depend on the appearance stream alone.
During inference, we use ResNet-50 features alone ($m_1$) for a fair comparison with competing approaches. 

\subsection{Key Step Localization}
Key Step Localization aims to localize each key step in a video. We follow the same evaluation protocol as prior work~\cite{bansal2022my,elhamifar2019unsupervised,kukleva2019unsupervised,shen2021learning,vidalmata2021joint}. The learned representations are first clustered and all frames are assigned to the various clusters. Since we work in an unsupervised setting, these frames are then matched to ground truth segments using the Hungarian algorithm~\cite{kuhn1955hungarian}. These mappings are then used to calculate the F-1 score and IoU scores used for comparison with prior works. F-1 score is the harmonic mean of precision and recall. Recall is computed as the ratio of number of frames having the correct key step prediction to the ground truth number of key frames across all key steps. Precision is the ratio of the number of correctly predicted frames and number of frames predicted as key steps. Following \cite{bansal2022my}, we report key-step averaged metric for the egocentric datasets and overall average for CrossTask and ProceL. We used the publicly available key step annotations provided by \cite{bansal2022my} for the egocentric datasets and use their evaluation code for a fair comparison. 

\begin{table*}
\begin{minipage}{1.0\linewidth}
\begin{center}
\caption{Procedure learning results on egocentric datasets. We see significant improvements using our approach on both F1 and IoU scores. We use the same backbone model as TC3I~\cite{bansal2022my} for a fair comparison. Note that unlike ~\cite{bansal2022my}, we do not finetune our backbones, or rely on video alignment. Our approach shows improvements using simple k-Means algorithm for clustering. All reported numbers for \ours{} are averaged over three runs.}
\label{table:egocentric_ksl}
{
\resizebox{0.8\linewidth}{!}{
\begin{tabular}{lccc|cccccccc}
\hline\noalign{\smallskip}
& Clustering  &Video  & Finetune& \multicolumn{2}{c}{\textit{CMU-MMAC}} & \multicolumn{2}{c}{\textit{EGTEA G.}} & 
\multicolumn{2}{c}{\textit{Meccano}} & \multicolumn{2}{c}{\textit{EPIC-Tents}}\\
Approach & Algorithm  & Alignment & Backbone & F1 & IoU & F1 & IoU  & F1 & IoU  & F1 & IoU \\
\noalign{\smallskip}
\hline
\noalign{\smallskip}
Random  & - & - & - & 15.7 & 5.9 & 15.3 & 4.6 & 13.4 & 5.3 & 14.1 & 6.5 \\
Uniform  & - & - & - & 18.4 & 6.1 & 20.1 & 6.6 & 16.2 & 6.7 & 16.2 & 7.9 \\
Bansal et al. ~\cite{bansal2022my}  & kMeans & \cmark & \cmark & 19.2 & 9.0 & 20.8 & 7.9 & 16.6 & 8.0 & 15.4 & 7.8 \\
Bansal et al. ~\cite{bansal2022my}  & PCM & \cmark & \cmark & 22.7 & \textbf{11.1} & 21.7 & 9.5 & 18.1 & 7.8 & 17.2 & 8.3 \\
\hline
\Ours{}  & kMeans & \xmark & \xmark  & \textbf{28.3} & \textbf{11.4} & \textbf{30.8} & \textbf{12.4} & \textbf{36.4} & \textbf{18.0} & \textbf{42.2} & \textbf{21.4} \\
\hline
\end{tabular}}
}
\end{center}
\end{minipage}
\end{table*}
\begin{table}
\begin{minipage}{1.0\linewidth}
\begin{center}
\caption{Procedure learning results on third person datasets. We compare with various state-of-the-art approaches and obtain consistent improvements. Unlike some prior works, we rely on appearance features alone at inference. We do not finetune our backbones, or rely on video alignment.}
\label{table:thirdperson_ksl}
{
\resizebox{\linewidth}{!}{
\begin{tabular}{l|cccccc}
\hline\noalign{\smallskip}
& \multicolumn{3}{c}{\textit{ProceL}} & \multicolumn{3}{c}{\textit{CrossTask}}\\
Approach & P & R & F1 & P  & R & F1 \\
\noalign{\smallskip}
\hline
\noalign{\smallskip}
Uniform   & 12.4 & 9.4 & 10.3 & 8.7 & 9.8 & 9.0 \\
Alyarc et al.~\cite{alayrac2016unsupervised}   & 12.3 & 3.7 & 5.5 & 6.8 & 3.4 & 4.5 \\
Kukleva et al.~\cite{kukleva2019unsupervised}   & 11.7 & 30.2 & 16.4 & 9.8 & 35.9 & 15.3 \\
Elhamifar et al.~\cite{elhamifar2020self}   & 9.5 & \textbf{26.7} & 14.0 & 10.1 & \textbf{41.6} & 16.3 \\
Fried et al.~\cite{fried2020learning}   &  - & - & - & - & 28.8 & - \\
Shen et al.~\cite{shen2021learning}   & 16.5 & 31.8 & 21.1 & 15.2 & 35.5 & 21.0 \\
Bansal et al.~\cite{bansal2022my}   & 20.7 & 22.6 & 21.6 & 22.8 & 22.5 & 22.6 \\
\hline
\Ours{}  &  \textbf{23.5} & \textbf{26.7} & \textbf{24.9}  & \textbf{26.2} & 25.8 & \textbf{25.9} \\
\hline
\end{tabular}}
}
\end{center}
\end{minipage}
\end{table}
We present our main results in \cref{table:egocentric_ksl} and \cref{table:thirdperson_ksl} for Egocentric and third person procedural datasets respectively. We note that we consistently outperform the prior works with often large margins. As demonstrated in the ~\cref{sec:ablations}, we can attribute the performance improvements to three key factors: ability to use long range dependencies due to use of pre-extracted features, use of multi-cue information during training, and refinement of the positive window for the contrastive learning. None of the prior works in procedure learning use long term temporal information in their features. We argue that these are essential to learning good features and is practically always available. Use of pre-extracted features makes this computationally tractable in our case. We would like to reiterate that for a fair comparison with prior works, we do \emph{not} use $m_2$ during evaluation and use the same feature extractor (ResNet-50) as Bansal et al.~\cite{bansal2022my}. 
\subsection{Phase classification and Kendall's Tau}
While Key-Step localization is the predominant way of evaluating approaches for procedure learning, our learned representations can be used for other tasks as well. Following prior work on video alignment \cite{haresh2021learning,liu2021learning}, we evaluate our learned representations for Phase Classification (PC). Phase classification calculates the per frame classification accuracy for fine-grained action recognition by training an SVM on the extracted per frame features of the training set. We show results on the IkeaAssembly dataset which is a dataset of people assembling Ikea furniture. The dataset depicts complex furniture assembly tasks performed by multiple people across views. The tasks are composed of multiple sub-tasks like flipping the table, attaching the leg etc. PennActions has videos of humans doing sports and exercises and are composed of multiple phases per action. We follow the same evaluation protocol as used in ~\cite{dwibedi2019temporal, haresh2021learningLong, liu2021learning}.  In addition, we also evaluate the Kendall's Tau (K.T) on the PennActions dataset\footnote{Since K.T assumes a strict montonic order of actions, we report it only for the PennActions dataset.} 
\begin{table*}
\begin{minipage}{1.0\linewidth}
\begin{center}
\caption{\small Proxy task results on IkeaASM and PennActions. Our feature encoder is able to adapt the off-the-shelf features for tasks like phase classification and Kendall's Tau (K.T). We show significant improvements over the state-of-the-art using the same backbone. Note that unlike prior approaches, we do not finetune our backbones. Use of motion information at inference leads to further improvements.}
\label{table:ikea_phasecls}
{
\resizebox{0.8\linewidth}{!}{
\begin{tabular}{lcc|cccccccccc}
\hline\noalign{\smallskip}
& & & \multicolumn{3}{c}{\textit{Ikea w/o Bgnd}} & \multicolumn{3}{c}{\textit{Ikea with Bgnd}} & 
\multicolumn{4}{c}{\textit{PennActions}}\\
Approach & Video & Finetune & \multicolumn{3}{c}{\textit{Phase Cls}} & \multicolumn{3}{c}{\textit{Phase Cls}} &
\multicolumn{3}{c}{\textit{Phase Cls}} & 
\multicolumn{1}{c}{\textit{K.T}} \\
 & Alignment & Backbone & 0.1 & 0.5 & 1.0 & 0.1 & 0.5 & 1.0 &  0.1 & 0.5 & 1.0 \\
\noalign{\smallskip}
\hline
\noalign{\smallskip}
Supervised  & \xmark & \cmark &  21.76 & 30.26 & 33.81& 20.74 & 25.61 & 31.92 &  67.10 & 82.78 & 86.05 & - \\
Random  & - & - & 17.89 & 17.89 & 17.89 & 17.03 & 17.41 & 17.61 &  44.18 & 46.19 & 46.81& -\\
ImageNet  & - & \xmark & 18.05 & 19.27 & 19.50 & 17.27 & 18.02 & 18.64 &  44.96 & 50.91 & 52.86 &- \\
SAL~\cite{misra2016shuffle} & \xmark & \cmark & 21.68 & 21.72 & 22.14 & 22.94 & 23.43 & 25.46 &  74.87 & 78.26 & 79.96  & 0.63 \\
TCN~\cite{sermanet2018time} & \xmark & \cmark & 25.17 & 25.70 & 26.80 & 22.51 & 25.47 & 25.88 &  81.99 & 83.67 & 84.04 & 0.73 \\
TCC~\cite{dwibedi2019temporal} & \cmark & \cmark & 24.74 & 25.22 & 26.46 & 22.70 & 25.04 & 25.63 &  81.26 & 83.35 & 84.45  & 0.74 \\
LAV~\cite{haresh2021learning} & \cmark & \cmark & 29.78 & 29.85 & 30.43 & 23.19 & 25.47 & 25.54 &  83.56 & 83.95 & 84.25  & 0.80 \\
VAVA~\cite{liu2021learning} & \cmark & \cmark & \textbf{31.66} & 33.79 & 32.91 & \textbf{29.12} & 29.95 & 29.10 &  \textbf{83.89} & 84.23 & 84.48 & 0.81  \\
\hline
\ours{} & \xmark & \xmark & 28.59 & \textbf{36.25} & \textbf{37.02} & 26.5 & \textbf{31.92} & \textbf{31.52} &  \textbf{83.34} & \textbf{85.06} & \textbf{85.5} & \textbf{0.91}\\

\hline
\end{tabular}}
}

\end{center}
\end{minipage}
\end{table*}
Table~\ref{table:ikea_phasecls} shows results of our approach on these metrics. The improved performance shows that our approach learns generalizable features for complex videos. We see that our \ours{} model which uses the same backbone as previous approaches (albeit without \textit{any} finetuning) outperforms the recent approach of VAVA on most metrics for both datasets. 

\subsection{KS Extraction}
Our key goal in this paper is to be able to sample a few key steps corresponding to the complex task for use in AR content creation. The sections above demonstrate the effectiveness of our learned features for related tasks of Key-Step Localization and Phase Classification. Since we do not have any labeled datasets to evaluate KS extraction, we show qualitative results instead. In Figure ~\ref{fig:km_qualitative}, we visualize the key steps extracted on the Ikea Assembly~\cite{ben2021ikea} dataset.
We visualize the extracted key steps for two randomly sampled videos for the tasks of assembling `Kallax Shelf Drawer' and `Lack Coffee Table' respectively.
We also manually map them to diagrams from the official IKEA manuals~\cite{ikea_manual}. We notice that the extracted key steps are semantically meaningful given the mapping. We did not include all steps from the manual since the curators of the dataset pre-assembled a few parts for ease of dataset collection. Note that what constitutes a key step can be very subjective and the number of steps needed to accomplish a task might depend on the application area. For example, a trainee trying to learn a new task might benefit from more steps than an an experienced person trying to transfer skills from a known task. For this visualization, we simply set the number of clusters $K=10$. We can also make use of clustering algorithms like \cite{finch} which automatically determine the number of clusters or even optimization-based algorithms to determine key steps given our learned representations. We include additional visualizations in the supplementary material. Our two-stage approach allows for this customization without having to retrain the model.

\subsection{Ablations and Analyses}
\label{sec:ablations}
In this section, we present ablations and analyses of our approach to better understand its working. We work with the Meccano bike assembly dataset for the following experiments unless otherwise stated. We include ablations for other datasets in the supplementary material.  
\\
\noindent\textbf{Use of On-Device sensor modalities:} Since we work with pre-extracted features, our approach easily allows use of on-device sensor modalities commonly available in many AR/VR headsets. These sensors often give complementary and distilled information about the scene which can be useful for the task at hand. For example, consider the task of assembling a toy bike with many parts. The visual stream sees a lot of clutter in front of the table. The gaze sensor returns very low dimensional information about which part of the bike the person is looking at and might be helpful. The Meccano dataset was captured using an Intel RealSense SR300 headset and a Pupil Core Gaze tracker mounted on the participant. The dataset includes synchronized RGB, Depth and Gaze signals. The results using these sensors are presented in ~\cref{table:sensors_mecanno}~(Use of on-device sensor modalities). We see that the use of on-device sensors improves the performance over model trained without any multi-cue training. Further, we see that Depth and gaze information is competitive to optical flow features. This is beneficial since for this dataset capture, these modalities comes for \emph{free}. Our approach enables plug-and-play use of additional modalities. 
\\
\begin{table}
\begin{center}
\caption{Benefits of using mutliple modalities. Our approach can easily be used with sensors often available on AR headsets. Use of multiple modalities at inference further improves performance.}
\label{table:sensors_mecanno}
\resizebox{1.0\linewidth}{!}{
\begin{tabular}{lllcc}
\hline\noalign{\smallskip}
Method & Training & Inference & F1 & IoU  \\
\hline
Random & - & - & 13.4 & 5.3 \\
Uniform & - & - & 16.2 & 6.7 \\
TC3I & RGB & RGB & 18.1 & 7.8 \\
\hline
\multicolumn{2}{l}{\textit{Ours without Multi-Cue training}}\\
\ours{} & RGB & RGB & 32.1 & 14.9 \\
\hline
\multicolumn{2}{l}{\textit{Use of RGB-derived modality}}\\
\ours{} & RGB+OF & RGB & \textbf{36.4} & \textbf{18.0} \\
\hline
\multicolumn{2}{l}{\textit{Use of on-device sensor modalities}}\\
\ours{} & RGB+Gaze & RGB & 35.4 & 17.0 \\
\ours{} & RGB+Depth & RGB & \textbf{36.6} & \textbf{18.1} \\
\hline
\multicolumn{2}{l}{\textit{Use of different modalities at inference}}\\
\ours{} & RGB+OF & RGB & 36.4 & \textbf{18.0} \\
\ours{} & RGB+OF & OF & 30.6 & 15.6 \\
\ours{} & RGB+OF & RGB+OF & \textbf{37.6} & \textbf{18.1} \\
\hline
\end{tabular}}
\end{center}
\vspace{-0.7cm}
\end{table}
\noindent\textbf{Use of MM information at inference:}
For a fair comparison, following prior works we use only appearance cue during inference. In the following experiment, we make use of multi-modal cues during inference as well. In Table~\ref{table:sensors_mecanno}~(Use of different modalities at inference), we show results with Appearance+Motion cues at inference. We simply concatenate the features from both modalities and use them for the downstream task. We see that even simple late fusion leads to improvements showing complementary information in the two streams. Note that while these are referred to as separate cues, RAFT features (optical flow) are derived from the RGB stream. 
\\

\begin{figure}
    \centering
    \includegraphics[width=0.7\linewidth]{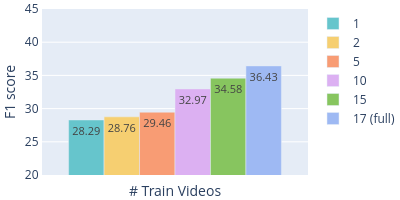}
    \caption{Amount of pretraining data. In this experiment we vary the amount of pre-training data that is used to train our encoder. Relying on off-the-shelf features lets us train our model to a good performance with even with a single video.}
    \label{fig:pretraining_meccano}
\end{figure}

\noindent\textbf{How much training data do we need?}
Our work follows a learning-by-observation style framework. Our goal is to be able to learn from a few recordings of the complex task by experts and hence a practical system should be able to work with little training data. Collecting unlabeled data for general tasks might be easy through scraping YouTube, but extracting a lot of procedural data through AR systems is non-trivial. In Figure ~\ref{fig:pretraining_meccano} we plot the F1 score as a function of training data on the Meccano dataset. We see that our approach gives good performance even with a single video. We hypothesize that this adaptability of the model to little training data is due to the use of pre-extracted features with a light-weight temporal module and use of long temporal sequences which gives the model lots of good signals during training. 
\\

\noindent\textbf{Ablation analyses on various losses:}
We quantitatively evaluate the effect of each our loss terms. In Table~\ref{table:loss_ablation} we show results on the Meccano dataset. We see that each of our proposed losses brings an improvement to the final performance. 
\\

\begin{table}
\begin{center}
\caption{Effect of different losses. In this experiment, we evaluate the effect of our different loss terms on the Meccano dataset. We see that our loss terms leads to performance improvements.}
\label{table:loss_ablation}
\resizebox{0.9\linewidth}{!}{\begin{tabular}{lcccc}
\hline\noalign{\smallskip}
Approach & Multi-Cue training & BMC2 & F-1 & IoU \\
\hline
TC3I & - & - & 18.1 & 7.8 \\
\hline
\ours{} & \xmark & \xmark & 32.1 & 14.9 \\
\ours{} & \xmark & \cmark & 33.1 & 15.8 \\
\ours{} & \cmark & \xmark & 34.2 & 16.6 \\
\ours{} & \cmark & \cmark & \textbf{36.4} & \textbf{18.0} \\
\hline
\end{tabular}}
\end{center}
\vspace{-0.3cm}
\end{table}

\noindent\textbf{Importance of long range temporal dependencies:}
Note that most previous approaches do not focus on the temporal aspect of the problem. 
We believe that temporal information is especially important for complex tasks since they often involve subtle and slow changes. Such changes may not be easily captured using a small temporal context used by some previous approaches. Prior works~\cite{haresh2021learning,liu2021learning} use very few sampled frames ($\sim$40) per video during training with little to no temporal context . This is sub-optimal since the videos from procedural datasets often contain 10000+ frames with sub-actions having long temporal spans (Figure 3~\cite{ben2021ikea}). Fine-tuning a backbone while dealing with long temporal contexts is computationally expensive. Use of pre-extracted features enables us to use large temporal context which helps in capturing more information from the video.  In~\cref{table:temporal_chunks}, we empirically verify the effect of long temporal contexts.  
\\

\begin{table}
\begin{center}
\caption{Effect of varying size of temporal context used for training. We see that the model achieves best performance at 1024 chunks, much higher than used by past approaches ($<$ 100).}
\label{table:temporal_chunks}

\begin{tabular}{lccccc}
\hline\noalign{\smallskip}
Window & 128 & 256 & 512 & 1024 & 2048\\
\hline
F1 & 33.04 & 36.08 & 34.91 & \textbf{36.43} & 34.88 \\
IoU & 15.64 & 17.75 & 17.01 & \textbf{18.01} & 16.38 \\
\hline
\end{tabular}
\end{center}
\vspace{-0.5cm}
\end{table}

\noindent\textbf{Effect of number of clusters}
\label{sec:num_clusters}
In ~\cref{table:num_clusters}, we evaluate our trained models for the Meccano and EPIC-Tents dataset by varying the number of clusters used during kMeans. We report the F1 score for this experiment. As noted by prior work, we obtain the best results at $K=7$. Our approach consistently outperforms the baselines.
\\
\begin{table}
\begin{minipage}{1.0\linewidth}
\begin{center}
\caption{Changing num clusters. We report the F1 scores for models trained on Meccano and EPIC-Tents dataset. We consistently outperform the state-of-the-art. }
\label{table:num_clusters}
{
\resizebox{0.8\linewidth}{!}{
\begin{tabular}{lcccc}
\hline\noalign{\smallskip}
Approach & K=7 & K=10 & K=12 & K=15 \\
\noalign{\smallskip}
\hline
\noalign{\smallskip}
\multicolumn{5}{c}{\textit{Meccano}}\\
Random  & 13.4 & 10.1 & 8.8 & 7.4\\
Uniform  & 16.2 & 13.2 & 11.8 & 11.2\\
Bansal et al. ~\cite{bansal2022my}  & 18.1 & 15.2 & 13.5 & 11.9\\
\Ours{}  & \textbf{36.4} & \textbf{31.2} & \textbf{27.5} & \textbf{25.6}\\
\hline
\multicolumn{5}{c}{\textit{EPIC-Tents}}\\
Random  & 14.1 & 10.6 & 9.1 & 8.3\\
Uniform  & 16.2 & 15.1 & 14.2 & 13.0\\
Bansal et al. ~\cite{bansal2022my}  & 17.2 & 11.1 & 12.1 & 9.5\\
\Ours{}  & \textbf{36.7} & \textbf{32.8} & \textbf{29.4} & \textbf{29.7}\\
\bottomrule
\end{tabular}}
}
\end{center}
\end{minipage}
\end{table}

\noindent\textbf{Alternate temporal encoders}
\label{sec:alternate_temporal_encoders}
In the main paper, we use a transformer model to capture long range dependencies. In this section, we use alternate temporal encoders like LSTMs and TCNs. In ~\cref{table:different_temporal_modules}, we compare these with a transformer based module. As noted by works in the past, transformers learn better features owing to the use of self-attention and thus are better suited for our problem. 
\\
\begin{table}
\begin{center}
\caption{Comparison between different temporal modeling types on Meccano dataset. We see that using a transformer gives us the best performance for both metrics. }
\label{table:different_temporal_modules}
\begin{tabular}{lcc}
\hline\noalign{\smallskip}
Temporal module & F1 & IoU \\
\hline
TCN & 32.5 & 14.8 \\
LSTM & 33.6 & 16.4 \\
Transformer & \textbf{36.4} & \textbf{18.0} \\
\hline
\end{tabular}
\end{center}
\end{table}

\noindent\textbf{Additional analyses and experiments:} Please refer to the supplementary material section for data and experiments including 1) comparison of features extracted from different backbones; 2) additional experiments and visualizations on key step extraction; 3) additional ablations and analyses; 4) dataset and implementation details; 5) Use of $>2$ modalities during training.

\section{Discussion and Conclusion}
\label{sec:conclusion}
We believe in the potential to use AR headsets to transform on-the-job training and guidance. Key step extraction from egocentric task videos will be critical to enabling AR-based task guidance at scale. In this work, we investigate this problem and offer substantial evidence that accounting for long-range temporal dependencies, use of multi-cue information during training and runtime, and refinement of the positive window for the contrastive learning are crucial to key step extraction. These findings are consistent with psychological research that advocates for detailed, multi-modal observation of human experts when designing training programs for novices \cite{gopher1994transfer, fadde2018training}. In future work, we hope to address several important related problems. Primary among these is collecting a detailed dataset of procedural task recordings from AR headsets. Also, incorporating recent advances in meta-learning and few-shot learning as ways to potentially improve key step extraction while reducing the amount of required training data is an important line of research. In short, the present work is only the beginning.
\section{Acknowledgements}
AS acknowledges support through a fellowship from JHU + Amazon Initiative for Interactive AI (AI2AI). RC acknowledges support from an ONR MURI grant N00014-20-1-2787. The authors are grateful to Bugra Tekin, Amol Ambardarekar, Susmija Reddy, Ketul Shah, Shlok Mishra and Aniket Roy for helpful comments and discussions. The authors would also like to thank Weizhe, Sanjay and Siddhant for answering questions regarding VAVA, LAV and EgoProceL respectively.

{\small
\bibliographystyle{ieee_fullname}
\bibliography{11_references}
}

\ifarxiv \clearpage \appendix
\label{sec:appendix}
\setlist[enumerate]{itemsep=0mm}

In this Supplementary material, we provide additional empirical studies, analyses and details. We list below the key sections.  

\begin{enumerate}
    \item \textbf{Feature extractor} : Details (\ref{sec:feature_extractor_details}) and results with additional modalities on Ikea dataset (\ref{sec:feature_extractor_additional_results}).
    \item \textbf{Bootstrapping} : Alternate variants (\ref{sec:bootstrap_variants})
    \item \textbf{Temporal encoder}: model details~(\ref{sec:model_details}), 
    Avoiding trivial solutions and other baselines~(\ref{sec:positional_encoding_analysis}).
    \item \textbf{Temporal sampling augmentation} : approach~(\ref{sec:temporal_sampling_augmentation}), effect of varying video extent~(\ref{sec:temporal_extent})
    \item \textbf{Miscellaneous analyses:} How long to train~(\ref{sec:how_long_to_train}), 
    Effect of inter-video alignment~(\ref{sec:video_align}), TC3I with multiple-cue training~(\ref{sec:tc3i_mm}), Additional results on loss ablation~(\ref{sec:loss_ablation_supplementary})
    \item \textbf{Key Step extraction:}
    Sampling variants ~(\ref{sec:alternative_sampling}), visualizations ~(\ref{sec:ks_visualization})
    \item \textbf{Practicality of the approach} : Time and resource requirements ~(\ref{sec:practicality})
    \item \textbf{Use of $>2$ modalities during training} ~(\ref{sec:n_modalities})
    \item \textbf{Additional details:} Datasets~(\ref{sec:dataset_details}), Additional information on KSL baselines ~(\ref{sec:baselines}), Evaluation protocols~(\ref{sec:evaluation_protocols}), Metrics~(\ref{sec:metric_details}), Hyperparameters~(Sec.~\ref{sec:hyperparam_and_implementation_details})
    \item \textbf{Overall flow} \ref{sec:overallflow}.
    \item \textbf{Limitations and future work}~(\ref{sec:limitations_and_future_work})
    \item \textbf{Negative Societal Impact}~(\ref{sec:negative_impact})
\end{enumerate}

\section{Feature extractor:}

\subsection{Details}
\label{sec:feature_extractor_details}
Our approach allows us to use off-the-shelf feature extractors instead of finetuning them. This is especially important
in a data-scarce and resource constrained setting. Note that unlike some prior works, we work with average pooled spatial features for quick training and lower storage requirements. 
\\
All results except Table 4 in the main paper used Resnet 50 as $m_1$ and RAFT/OF features as $m_2$. 

\noindent\textbf{Res50:} Features are extracted from the conv5c layer of a ResNet-50~\cite{he2016deep} backbone. We use ImageNet pretrained models following the prior
works~\cite{dwibedi2019temporal,haresh2021learning,liu2021learning}.

\noindent\textbf{RAFT/OF:} RAFT~\cite{teed2020raft} is a model trained for extracting optical flow from a pair of images. We extract motion features from the pre-trained feature encoder (last recurrent update). 

For experiments in Table 4:

\noindent\textbf{Gaze:} For sensor derived modalities, we consider the on-device hardware/software block as our feature extractor. For example, in the Meccano dataset, gaze data is made available as the x \& y location of the gaze mapped to the image coordinates along with a confidence score. Since gaze is recorded at 200Hz compared to 12Hz for the visual stream, we associate each from of the video to 16 (~$200/12$) frames thus giving us a $3 \times 16 = 48$ dimensional gaze vector corresponding to each frame.

\noindent\textbf{Depth:} Depth maps are encoded using a pretrained ResNet-50 and features extracted from the Conv5c layers are used for training. 

\subsection{Additional Results}
\label{sec:feature_extractor_additional_results}
Separating feature extractor step from the SSL lets us use any available off-the-shelf backbone. We next explore alternate feature extractor (Pose) for the IkeaASM dataset ~\cref{table:feature_ext_Ikea}.

\noindent\textbf{Pose:} We extract human pose coordinates using OpenPose~\cite{cao2017realtime}. The pose coordinates of each joint are stacked and used as input to the temporal encoder. We also experimented with using features from a pre-trained pose-based action recognition model. Specifically, we used features extracted from a FineGym and NTU pretrained 
PoseC3D~\cite{duan2022revisiting} model. We empirically found that this approach gave us a similar performance as using pose coordinates and thus we use coordinates for our pose-based experiments in the paper. We use the pose modality for experiments with the Ikea dataset due to the nature of videos (third person, static camera). For the unconstrained videos of the other datasets, we make use of our RAFT-based feature extractor. 

In ~\cref{table:feature_ext_Ikea} we compare the performance of different feature extractors for the task of phase classification on the Ikea dataset. During inference, for a fair comparison to the prior works we use the appearance feature alone unless specified otherwise. We also experiment with use of multiple modalities at inference (like Table 4 main paper) and use a simple concatenation of features (e.g. Res50 + RAFT). We see that use of multiple modalities at inference leads to additional gains. 

\begin{table}
\begin{center}
\caption{Comparison between different feature extractors for Ikea dataset. We use multi-cue features for training and evaluate on single/multi-cue features for inference.}
\label{table:feature_ext_Ikea}
\begin{tabular}{llc}
\hline\noalign{\smallskip}
Training & Inference & Phase CLS 1.0\\
\hline
Res50+Pose & Res50 & 30.6\\
Res50+Pose & Pose & 30.3 \\
Res50+Pose & Res50+RAFT & 31.9 \\
\hline
Res50+RAFT & Res50 & 31.5\\
Res50+RAFT & RAFT & 30.3\\
Res50+RAFT & Res50+RAFT & 31.7\\
\hline

\end{tabular}
\end{center}
\end{table}

\section{Bootstrapping}

\subsection{Alternate variants}
\label{sec:bootstrap_variants}
In the main paper, we discussed our proposed technique of improving the set of positives by using raw features to bootstrap the temporal windows. We noted (Section 3.3, main paper) that our final window used $\tilde{\mathcal{W}}$ was a union of the $\sigma$-window and the one obtained through bootstrapping. We refer to this approach as `Union-Window'. In this section we show our results with some alternate variants that we tried. 

\noindent\textbf{Only the sampled window:} In this approach we discard the $\sigma$-window for the loss computation and use only the bootstrapped window $\mathcal{W}'$. 

\noindent\textbf{Union Window, only modifying negative set} : Here, the positives come from the $\sigma$-window but instead of using the complement as negatives, we also remove the potential false negatives obtained using $\tilde{\mathcal{W}}$.

\begin{table}
\begin{center}
\caption{Comparing bootstrap variants on Meccano dataset. We compare variants for using the bootstrapped window. We notice that using the union window for defining both positives and negatives leads to the best results. }
\label{table:bootstrap_variants}
\resizebox{0.8\linewidth}{!}{\begin{tabular}{lcc}
\hline\noalign{\smallskip}
Variant & F-1 & IoU \\
\hline
No bootstrap & 32.1 & 14.9 \\
\hline
Only sampled window & 34.4 & 16.3\\
Union Window for negative & 34.6 & 16.3\\
Union Window for positive and negative & \textbf{36.4} & \textbf{18.0} \\
\hline
\end{tabular}}
\end{center}
\vspace{-0.3cm}
\end{table}
In ~\cref{table:bootstrap_variants}, we compare these three variants. We see that our approach of using the union window for both positive and negative set shows the best performance. Using the sampled window performs since it doesn't necessarily impose the temporal constraints for the loss function and ignore the $\sigma$-window. Use of the union window only to generate the negative set doesn't modify the false negatives as positives and shows poor performance.

\section{Temporal Encoder}

\subsection{Model details}
\label{sec:model_details}
We train a separate temporal encoder per modality. Each temporal encoder is a multi-headed transformer model. The input to the temporal encoder are raw features ${p}_{t}^{m_i} \in \mathbb{R}^{D_i}$. We use a two layer vanilla transformer with two heads. We do not use causal masks. The temporally adapted sequence ($\tilde{q}_{t}^{m_i} \in \mathbb{R}^D$) is then passed through a two layer MLP to obtain features ($q_{t}^{m_i} \in \mathbb{R}^D$) used for computing the loss. $D$ is set to 128. 

\subsection{Analysis : Positional Encoding}
\label{sec:positional_encoding_analysis}
In ~\cref{table:analysis_positional_encoding}, we analyse different ways of encoding temporal information into the model. Since transformers on its own does not have any information about the relative position of various steps in the temporal dimensions, we add a sinusoidal position encoding~\cite{vaswani2017attention} to the input sequence before adapting it with the transformer. Since our model will be subsequently trained with losses that enforce temporal consistency, any learnable embedding before adding position encoding might learn to ignore the raw features (~\cref{table:analysis_positional_encoding} \emph{STEPs w/ MLP before PosEnc}). Hence, we add this embedding to the raw feature directly to avoid a collapse of the learned representations. Note that removing the positional encoding performs worse (\emph{STEPs w/o PosEnc}) which shows the importance of encoding temporal information into the feature sequence. We see the same trend when using a loss with and without bootstrapping. Our final approach with bootstrapping outperforms other baselines. 
While this baseline still trains a temporal encoder, another potential baseline is to use the raw features directly for evaluation. This baseline (\emph{Raw features}) performs much worse showing the importance of adapting the features through our training loss. 
\begin{table}
\begin{center}
\caption{Comparing different variants of positional encoding on Meccano dataset. We see the benefits of using a positional encoding during training to impart ordering information. The positional encoding is added directly to raw features thus precluding the model from learning trivial embeddings. We see similar trends for model w \& w/o the bootstrapping window. }
\label{table:analysis_positional_encoding}
\begin{tabular}{lcc}
\hline\noalign{\smallskip}
Approach & F1 & IoU \\
\hline
TC3I~\cite{bansal2022my} & 18.1 & 7.8 \\
\hline
Raw Features & 20.1 & 8.6 \\
\hline
\textit{MC2 loss}\\
\ours{} w/o PosEnc & 27.7 & 12.7 \\
\ours{} w/ MLP before PosEnc & 28.8 & 12.9 \\
\ours{} & 34.2 & 16.6 \\
\hline
\textit{BMC2 loss}\\
\ours{} w/o PosEnc & 32.2 & 14.7 \\
\ours{} w/ MLP before PosEnc & 32.8 & 15.3 \\
\ours{} & \textbf{36.4} & \textbf{18.0} \\
\hline
\end{tabular}
\end{center}
\end{table}. 

\section{Temporal Sampling augmentation}

\subsection{Sampling approach}
\label{sec:temporal_sampling_augmentation}
Most works in Video-SSL work with short clips. We work with procedures which are minutes long, often with steps that might repeat in the future. Our approach trains a light-weight temporal encoder on pre-extracted features thus enabling use of long-range dependencies. Note that since our approach works with features instead of raw videos/images, we cannot employ any input-space augmentations like random resize crop, color jitters, blurring etc during SSL training. We apply a temporal sampling augmentation but on features and do not generate multiple `views' of the video. Instead, we extract positives and negatives from a single sample of the video. Our sampling strategy during training borrows ideas from TSN~\cite{wang2018temporal} but applies it to feature sequences. Given the complex video, we first randomly select a start $t_{start}$ and end frame $t_{end}$. These are sampled based on a hyperparameter of how much temporal extent of the video we want to cover during training. Next, the selected extent is uniformly divided into $N$ chunks. A feature frame from each chunk is randomly sampled which is then used in the subsequent model. Note that the corresponding time stamp of the sampled frame is utilized in positional encoding \cref{sec:positional_encoding_analysis}. The flow is presented in ~\cref{fig:temporal_sampling_viz}.
\begin{figure*}
    \centering
    \includegraphics[width=0.9\linewidth]{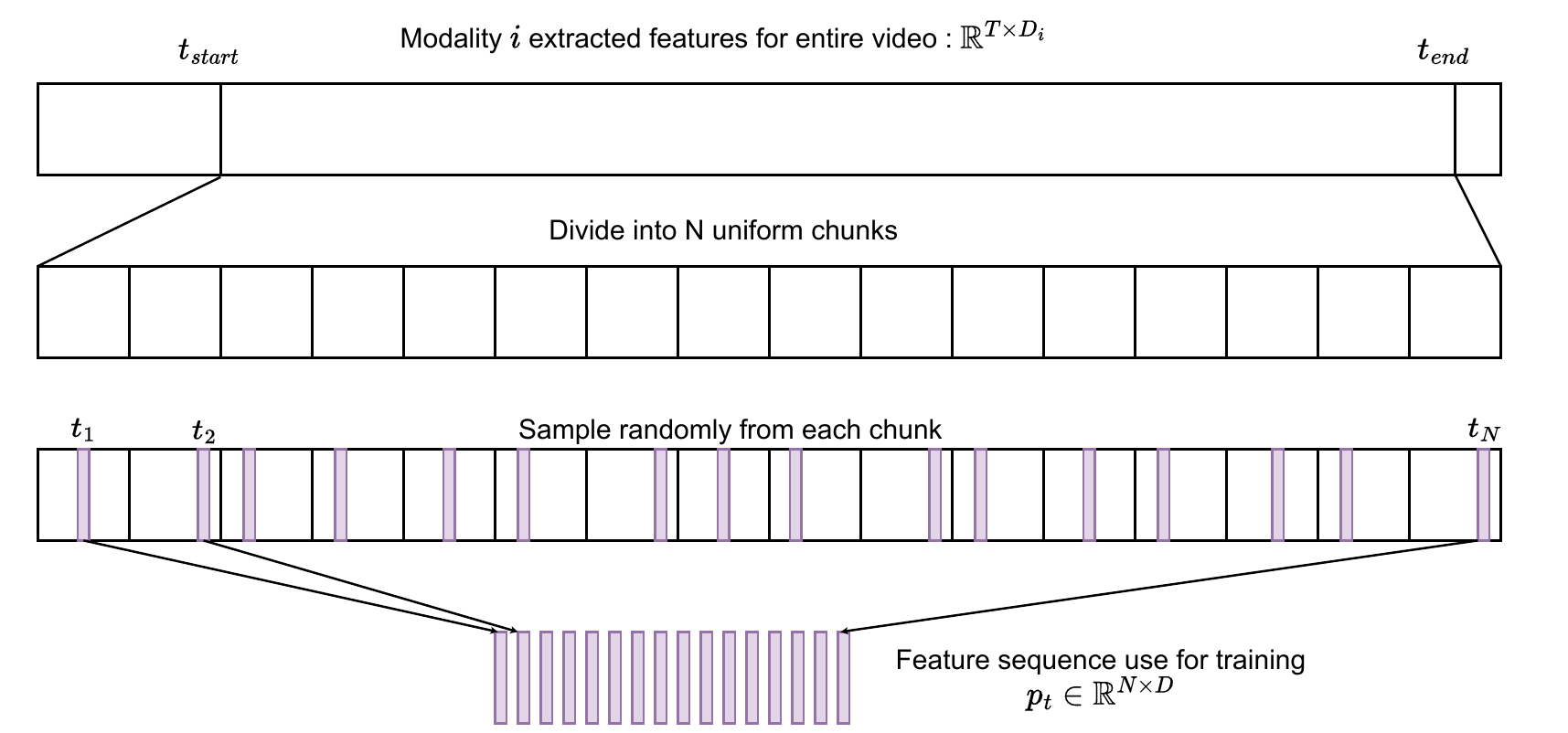}
    \caption{Temporal sampling. We illustrate the sampling applied during training. We first set a temporal extent. This is divided into N equal sized chunks. A feature frame is sampled at random from each chunk which is used to contruct the input sequence. }
    \label{fig:temporal_sampling_viz}
\end{figure*}

\subsection{Effect of temporal extent}
\label{sec:temporal_extent}
Given a temporal extent $\beta$, $t_{start}$ and $t_{end}$ are sampled from $[1, \cdots, T]$ such that $t_{end} - t_{start} \simeq \beta T$
All experiments in the main paper use $\beta=1$ which implies $t_{start}=1$ and $t_{end}=T$. In ~\cref{table:temporal_extent}, we experiment with different temporal extents and the effect on performance on the Meccano dataset. We note that $\beta=1$ gives the best performance. 
\begin{table}
\begin{center}
\caption{Comparing temporal extents. We sample our features from varying temporal extents of the video. $\beta$ denotes the extent of the whole video used for sampling. Using $\beta=1$ shows the best performance.}
\label{table:temporal_extent}
\begin{tabular}{lcc}
\hline\noalign{\smallskip}
Temporal Extent ($\beta$) & F-1 & IoU \\
\hline
0.2 & 24.7 & 10.7 \\
0.4 & 26.7 & 11.2\\
0.6 & 30.6 & 13.9\\
0.8 & 35.1 & 17.0\\
1.0 & \textbf{36.4} & \textbf{18.0}\\
\hline
\end{tabular}
\end{center}
\vspace{-0.3cm}
\end{table}

\section{Miscellaneous Analyses}

\subsection{How long to train?}
\label{sec:how_long_to_train}
We train all Meccano models for 300 epochs~(\cref{sec:hyperparam_and_implementation_details}). In this section we explore the effect of longer training of our models. We train our model for longer on the Meccano dataset. With a learning rate drop at 400 epochs, the model obtains F1 score of 37.8 and IoU of 18.25 at 500 epochs which shows that the model gradually improves even beyond 300 epochs but the marginal gains are low as is often observed in self supervised learning.

\subsection{Effect of inter-video alignment}
\label{sec:video_align}
Our approach shows strong performance compared to prior works without using any alignment-based loss. This lets us train even on a single video. An additional video-alignment objective is complementary to our contributions. Upon including inter-video alignment loss (soft-DTW~\cite{haresh2021learning}), we notice that the IoU improves by 0.7 with no improvement in F1 score. 

\subsection{TC3I with multiple modalities}
\label{sec:tc3i_mm}
Use of multiple modalities during training can benefit other approaches too. We experiment with including multi-cue training with the publicly available implementation of TC3I. Using RAFT (Optical flow) cues during training improves IoU on Meccano from 7.8 to 8.6. We notice that \ours{} still faares much better since it can capture long range temporal dependencies and benefit from bootstrapping of raw features. 

\subsection{Additional results on loss ablation}
\label{sec:loss_ablation_supplementary}
\begin{table*}
\begin{minipage}{1.0\linewidth}
\begin{center}
\caption{Effect of different losses. In this experiment, we evaluate the effect of our different loss terms on the four egocentric datasets. We see that our loss terms leads competitive performance compared to the baseline.}
\label{table:loss_ablation_appendix}
{
\resizebox{0.8\linewidth}{!}{
\begin{tabular}{lcc|cccccccc}
\hline\noalign{\smallskip}
& & & \multicolumn{2}{c}{\textit{CMU-MMAC}} & \multicolumn{2}{c}{\textit{EGTEA G.}} & 
\multicolumn{2}{c}{\textit{Meccano}} & \multicolumn{2}{c}{\textit{EPIC-Tents}}\\
Approach & Multi-Cue training & BMC2 & F1 & IoU & F1 & IoU  & F1 & IoU  & F1 & IoU \\
\noalign{\smallskip}
\hline
\noalign{\smallskip}
Random  & - & - & 15.7 & 5.9 & 15.3 & 4.6 & 13.4 & 5.3 & 14.1 & 6.5 \\
Uniform  & - & - & 18.4 & 6.1 & 20.1 & 6.6 & 16.2 & 6.7 & 16.2 & 7.9 \\
Bansal et al. ~\cite{bansal2022my} & & & 22.7 & \textbf{11.1} & 21.7 & 9.5 & 18.1 & 7.8 & 17.2 & 8.3 \\
\ours{} & \xmark & \xmark & 26.2 & 10.7 & 29.5 & 12.2 & 29.9 & 14.1 & 37.1 & 20.0\\
\ours{} & \xmark & \cmark & 25.0 & 9.6 & 29.6 & \textbf{12.4} & 33.1 & 15.8 & 37.7 & 19.0 \\
\ours{} & \cmark & \xmark & 27.7 & 11.0 & \textbf{30.8} & \textbf{12.4} &32.0 & 15.3 & 41.2 & \textbf{21.9}\\
\ours{} & \cmark & \cmark & \textbf{28.3} & \textbf{11.4} & 29.0 & 11.6 & \textbf{36.4} & \textbf{18.0} & \textbf{42.2} & \textbf{21.4} \\
\hline
\end{tabular}}
}
\end{center}
\end{minipage}
\end{table*}
In Table~\ref{table:loss_ablation_appendix}, we present the effect of various components of our approach on four egocentric datasets. We see that our proposed losses brings an improvement to the final performance in most cases. 

\section{Key Step extraction}

\subsection{Using alternate sampling approach for key step extraction}
\label{sec:alternative_sampling}
Separating representation learning from generation allows us to easily use alternate approaches to extract key steps. We experiment with the approach ILS-SUMM~\cite{shemer2021ils} which implements iterative local search for unsupervised video summarization. They recover an optimal summary by solving a constraint optimization problem on the total summary duration. The approach leads to plausible key steps (\cref{fig:ils_summary}) which show the effectiveness of our learning algorithm. We use our proposed clustering \& sampling steps since they provide more user control for key step extraction focused on task-based videos while ILS-SUMM provides fewer controls. For example, the k-Means clustering algorithm can easily be swapped with an alternative algorithm like FINCH~\cite{finch} which can automatically detect the number of clusters. In Figure ~\ref{fig:finch_summary}, we show results by swapping the k-Means in Algorithm 1 with FINCH. For an easy comparison with other presented results, we sort and visualize 10 key steps which have the least distance to the cluster centers. We see that use of this alternate clustering algorithm gives plausible key-steps.  

The example shows that our algorithm allows for easy customization based on user requirements. 

\begin{figure*}
    \centering
    \includegraphics[width=\linewidth]{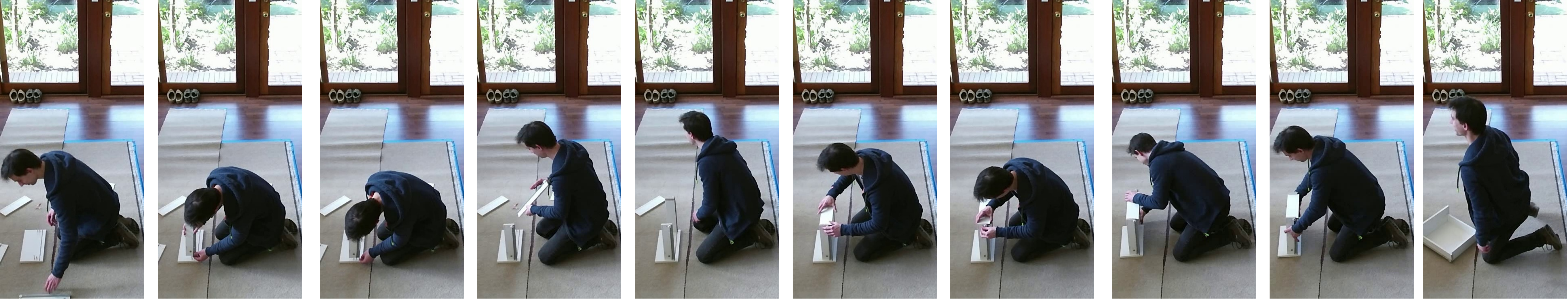}
    \caption{Alternative sampling approaches. We use our extracted features with ILS-SUMM to extract a 10 step summary of a person assembling a Kallax Shelf Drawer. The approach leads to plausible key steps showing the efficacy of our features. We use our proposed clustering \& sampling steps since they provide much more user control for key step extraction focused on task-based videos.}
    \label{fig:ils_summary}
\end{figure*}

\begin{figure*}
    \centering
    \includegraphics[width=\linewidth]{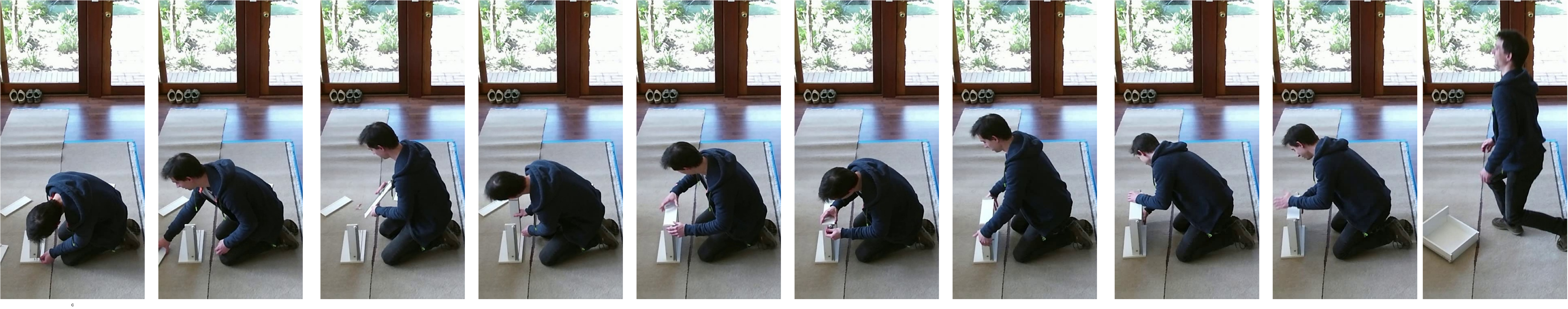}
    \caption{Alternative clustering algorithm. We replace k-Means with FINCH~\cite{finch} and visualize the top-10 key steps based on distance to the cluster centers. While this clustering approach seems to miss a step, we see that the results are still plausible. Our algorithm allows for easy customization based on user requirements.}
    \label{fig:finch_summary}
\end{figure*}

\subsection{Alternative KSE metrics}
Prior works like \cite{kaushal2019demystifying} proposed metrics for evaluating video summarization systems. We explore the evaluation of representation and diversity metrics from~\cite{kaushal2019demystifying} for the extracted key-steps on Meccano dataset. We compare these scores to key steps extracted using TC3I's features for the same number of steps extracted. We notice that our approach leads to a better representation score (53.1 vs 48.2) and higher min-disparity (diversity) score (6.52 vs 4.54). Since we are dealing with procedural videos rather than generic home-videos, we believe that KSL is a more apt proxy for KSE. Following prior works, we evaluate on KSL metric in this paper.  

\subsection{Additional KS visualizations}
\label{sec:ks_visualization}
\begin{figure*}
    \centering
    \includegraphics[width=\linewidth]{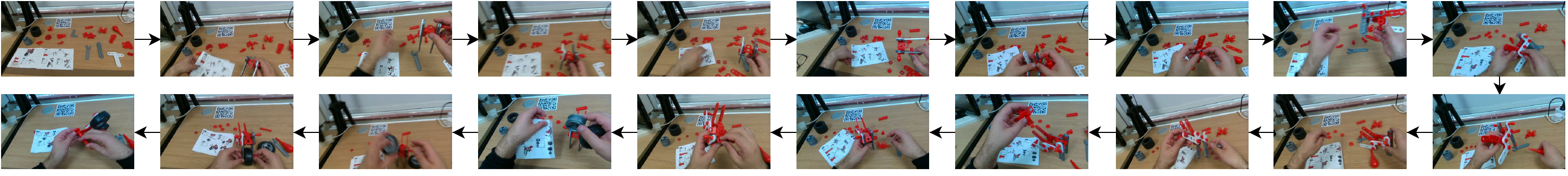}
    \caption{KS Visualization for Assembing a Toy Bike (Meccano dataset). We use the features extracted using our approach and extract a 20 step summary of assembling a toy model of a motorbike. We use the cluster and sample approach to generate these. We notice that the keys steps are plausible.}
    \label{fig:meccano_ks_viz}
\end{figure*}
\begin{figure*}
    \centering
    \includegraphics[width=\linewidth]{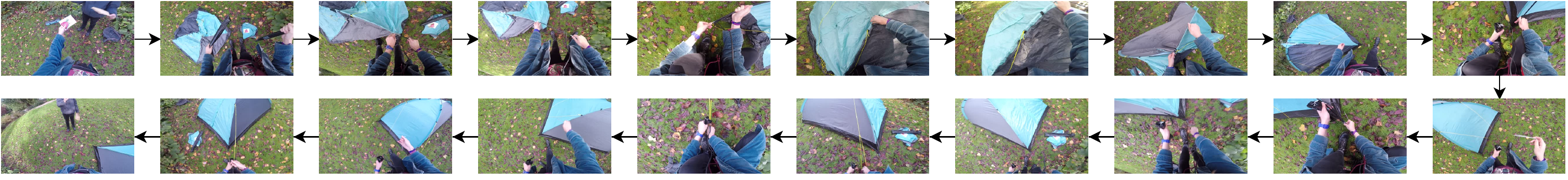}
    \caption{KS Visualization for Assembing a tent (EPIC-Tents dataset). We use the features extracted using our approach and extract a 20 step summary of assembling a tent. We use the cluster and sample approach to generate these. We notice that the keys steps are plausible.}
    \label{fig:epic_ks_viz}
\end{figure*}
In ~\cref{fig:meccano_ks_viz} and ~\cref{fig:epic_ks_viz}, we visualize the extracted key steps for a video of Meccano and EPIC-Tents respectively. We observe that the extracted key steps are plausible for the task of assembling a toy bike and assembling a tent respectively. 

\section{Practicality of the approach}
\label{sec:practicality}
The use of off-the-shelf feature extraction without fine-tuning enables fast transformer encoder training. We can train a model on 17 Meccano videos for 300 epochs in under 6 minutes on a single Nvidia A5000 GPU. While we use a 24GB GPU, for our set of hyperparameters, the training run needs only $\sim$ 3.5GB GPU memory. We use a single GPU for all our experiments. Our small memory footprint (1.8M parameters for our model compared to 23M in the Res50 encoder alone) along with the use of off-the-shelf features allows training with large temporal extents unlike approaches which rely on finetuning. 
 
We note that precomputing of features is also quite fast. We use standard pipelines for feature extraction. For example, using a single GPU, it takes less than 15 minutes to extract Resnet50 appearance features for \emph{all} frames ($> 300k$) of Meccano dataset. Further, some cues like gaze do not need any feature extraction and can be used in the raw form.

\section{Use of multiple modalities during training}
\label{sec:n_modalities}
In the main paper we use 2 modalities for all of our experiments. In this section, we train our approach using $>2$ modalities while performing inference on the RGB/appearance modality alone for an easy comparison. We present our results in the ~\cref{table:n_modalities}. We observe that use of additional modalities benefits from longer training and we train these models for 700 epochs with a LR drop at 600. For simplicity, in these experiments, we set 
  \[
    \lambda_{uv} = \left\{\begin{array}{lr}
        1, & \text{for } u = v\\
        1, & \text{for } u = \texttt{RGB}, v \neq \texttt{RGB}\\
        0, & \text{otherwise} 
        \end{array}\right\}
  \]
We see that using additional modalities during training consistently improves performance. In particular, we note that depth and optical flow are especially helpful. Finally, using all four modalities during training gives further improvements and gives the best results. 
\begin{table}
\begin{center}
\caption{Training with $N>2$ modalities}
\label{table:n_modalities}
\resizebox{1.0\linewidth}{!}{
\begin{tabular}{lllcc}
\hline\noalign{\smallskip}
Method & Training & Inference & F1 & IoU  \\
\hline
\multicolumn{2}{c}{\textit{Training using $3$ modalities}}\\
\ours{} & RGB+OF+Gaze & RGB & 37.8 & 18.7 \\
\ours{} & RGB+OF+Depth & RGB & 38.7 & 19.0 \\
\ours{} & RGB+Gaze+Depth & RGB & 37.8 & 18.7 \\
\multicolumn{2}{c}{\textit{Training using $4$ modalities}}\\
\ours{} & RGB+OF+Gaze+Depth & RGB & \textbf{40.2} & \textbf{19.8} \\
\hline
\end{tabular}}
\end{center}
\end{table}

\section{Additional details:}

\subsection{Dataset details}
\label{sec:dataset_details}
\subsubsection{Egocentric procedure learning datasets}
We use four egocentric datasets to validate our approach. We use the publicly available annotations from the authors of CnC~\cite{bansal2022my} for our experiments on these datasets. 

\noindent\textbf{CMU-Kitchens/MMAC~\cite{de2009guide}:} This dataset contains recording of  subjects performing the tasks involved in cooking and food preparation. A kitchen was built and twenty-five subjects were recorded to cook five different recipes: brownies, pizza, sandwich, salad, and scrambled eggs.

\noindent\textbf{EGTEA Gaze+~\cite{li2018eye}:} Extended GTEA Gaze+ is a large-scale dataset with modalities like visual, gaze tracking, audio etc. It consists of activities from 86 unique sessions of 32 subjects. The dataset includes seven cooking recipes : Continental Breakfast, Pizza, Bacon and eggs, Cheese burger, Greek salad, Pasta salad and Turkey sandwich.

\noindent\textbf{Meccano Dataset~\cite{ragusa2021meccano}:} Meccano is a multimodal egocentric video dataset of people assembling a toy bike. The dataset is captured with multiple modalities like RGB videos, depth videos and gaze signals. We use this dataset for most ablations and analyses.

\noindent\textbf{EPIC-Tents Dataset~\cite{jang2019epic}:} This is an outdoor egocentric video dataset of people assembling a camping tent. The dataset is collected from 29 participants. 

\subsubsection{Third-person procedure learning datasets}
\noindent\textbf{ProceL Dataset~\cite{elhamifar2019unsupervised}}: consists of videos from 12 procedures like replacing iPhone battery, setting up Chromecast etc. The dataset consists of about 720 videos with 8 key-steps on average. The videos are obtained from YouTube. We use the same experimental setup and features as used in ~\cite{bansal2022my} following their official implementation.

\noindent\textbf{Crosstask Dataset~\cite{zhukov2020learning}}: is an instructional video dataset with 18 primary tasks. The dataset consists of 2750 videos with 7 keys-steps on average. We use the same experimental setup and features as used in ~\cite{bansal2022my} following their official implementation.

\noindent\textbf{Ikea-Assembly Dataset~\cite{ben2021ikea}}: comprises of complex furniture assembly tasks performed by multiple people across views. The challenging tasks are composed of multiple sub-tasks like flipping the table, attaching the leg etc. While captured in a controlled setting, the dataset consists of large temporal variations and includes sections where no activity takes place. This dataset closely mimics the practical AR/VR scenario that we discuss in the introduction of the main paper. We follow the same experimental setting as the prior work~\cite{haresh2021learning,liu2021learning} for a fair comparison. The top-view was used for all experiments. Kallax Shelf-Drawer assembly task split was used unless otherwise specified.

\subsubsection{Other datasets}
\noindent\textbf{PennActions Dataset~\cite{zhang2013actemes}}: The dataset consists of 13 action categories from the PennAction dataset as used by authors in \cite{liu2021learning,haresh2021learning,dwibedi2019temporal}. The actions are composed of humans doing sports and exercises and are composed of 2-6 phases per action. While this is not strictly a procedure learning datasets, we apply our method to this approach to show the wide applicability.

\subsection{More information on KSL baselines}
\label{sec:baselines}

\noindent\textbf{Random:} Following ~\cite{bansal2022my}, we assign a random label to each frame from a uniform distribution with K values. These K values represent K steps. 

\noindent\textbf{Uniform:} Here, we uniformly divide the video into K chunks and assign a unique label to each chunk. We found this to be a stronger baseline than `Random'. 

\noindent\textbf{Bansal et al~\cite{bansal2022my}:} Here we directly report the best results obtained by the most relevant prior work. This approach trains an embedding network using a combination of losses. The frames for evaluation videos are then assigned to K labels using a ProCut module which relies on soft-clustering. We report their best results for each compared dataset.

\subsection{Evaluation protocols}
\label{sec:evaluation_protocols}
We follow the same evaluation protocols as prior works for a fair comparison. For CMU-MMAC, EGTEA-Gaze+, EPIC-Tents, Meccano, ProceL and CrossTask, we follow the evaluation protocol used in CnC~\cite{bansal2022my}. Videos were sampled at FPS/2. We use the improved metric proposed in by authors of ~\cite{bansal2022my} which evaluates the IoU and F1 score per-key step and averages them. Closely following ~\cite{bansal2022my}, this modified protocol is not used for CrossTask and ProceL datasets where standard protocol~\cite{elhamifar2019unsupervised,elhamifar2020self,kukleva2019unsupervised,shen2021learning} of calculating it for all steps together is used. For experiments on IkeaASM and PennActions, we closely follow the protocols laid out in VAVA~\cite{liu2021learning} and LAV~\cite{haresh2021learning}. For IkeaASM, evaluation was performed on frames sampled at 8FPS, while for PennActions, they are sampled at original FPS during evaluation.

\subsection{Metric details}
\label{sec:metric_details}
Here we present additional details on the various metrics used to evaluate our models. For all metrics, a higher score implies a better performance. 

\noindent\textbf{Key-step localization:} We follow the same experimental setup as ~\cite{bansal2022my}. We first find a one-to-one matching between steps in the ground truth and clustering predictions from our method using the Hungarian algorithm. Recall is computed as the ratio of number of frames having the correct key step prediction to the ground truth number of key frames across all key steps. Precision is the ratio of the number of correctly predicted frames and number of frames predicted as key steps. F-1 score is the harmonic mean of recall and precision.

\noindent\textbf{Phase Classification:} We calculate the average per-frame phase classification accuracy obtained by training an SVM on the phase labels on our temporally adapted per-frame features. Following prior works~\cite{dwibedi2019temporal,haresh2021learning,liu2021learning}, we evaluate the model on varying amounts of labels used for SVM training (0.1, 0.5 and 1.0). 

\noindent\textbf{Kendall`s Tau:} This is a statistical measure which is used to determine the temporal alignment between two sequences. Since this metric assumes a strict monotonic order of actions and we report it only for the PennActions dataset. Given two videos, we first sample a pair of frame features $q_i,q_j$ from the first video and retrieve the corresponding closest features in the second video of the same task $q_{i'},q_{j'}$. The frame indices $i,j,i',j'$ as concordant if $(i-j)(i'-j')>0$ and discordant other wise.  Kendall's Tau is calculated as 
\begin{equation}
    K.T = \frac{\# \text{concordant pairs} - \text{\# discordant pairs}}{\binom{n}{2}}
\end{equation}
Please refer to \cite{dwibedi2019temporal} for further details.

\subsection{Hyperparameter details}
\label{sec:hyperparam_and_implementation_details}

We use PyTorch for training our models. 
In ~\cref{table:hparams}, we list the common per-dataset hyperparameters used for our experiments in Tables 1, 2 \& 3 of the main paper. Following are some additional training and implementation details. 

\begin{table}
\begin{center}
\caption{Common hyperparameters used for training STEPs models}
\label{table:hparams}
\begin{tabular}{lc}
\hline\noalign{\smallskip}
Hyperparameter & Value \\
\hline
Clustering Algorithm & k-Means\\
$\#$ Clusters & 7 \\
$\lambda_{uu}$ & 1 \\
$\lambda_{uv}$ & 1 \\
margin $\zeta$ & 2.0 \\
Learning rate & 1E-3 \\
Train modalities & Res50 + RAFT-OF \\
Inference modalities & Res50 \\
Number of heads & 2 \\
Dimension of hidden layers & 128 \\
Number of transformer layers & 2 \\
Temporal extent $\beta$ & 1.0 \\
Optimizer & Adam \\
\hline
\end{tabular}
\end{center}
\end{table}

\noindent\textbf{Batch size:} We use a batch size of 4 for all datasets except CrossTask, ProceL and PennActions which use a batch size of 16. 
\\
\noindent\textbf{Temporal window size:} We use a window size of 10s for all datasets except PennActions. Due to the small temporal lengths in that dataset, we instead use a window size of 4 frames for that dataset.
\\
\noindent\textbf{Epochs:} We train models on Meccano and EGTEA for 300 epochs and on EPIC-tents and CMU-Kitchens for 150 epochs. Models on CrossTask and ProceL were trained for 100 epochs due to large size of the dataset. Models on Ikea were trained for 300 epochs while those for PennActions were trained for 400 epochs. 
\\
\noindent\textbf{Modality for bootstrapping:} We obtain the best results when we use RGB modality for bootstrapping on Meccano, EPIC, and ProceL datasets. Raw RAFT features were used for CrossTask while a concatenation of RGB and RAFT were used for CMU Kitchens, EGTEA, Ikea and PennActions.
\\
\noindent\textbf{Number of chunks:} We use 1024 chunks for all datasets except CrossTask, ProceL and PennActions. Both CrossTask and ProceL use 512 chunks. Due to the very short lengths of videos in PennActions, we determine number of chunks as $0.8$ of average video length in that dataset. 
\\
\noindent\textbf{Average over runs:} Since most of the datasets we work with are small in size, we report all results for Meccano, EPIC-Tent, CMU-Kitchens, EGTEA and Ikea as average of 3 training runs from random seeds.

\noindent\textbf{Key step extraction details}
For our key step visualizations, we first cluster the video features. For k-Means, we set number of clusters as the number of phases in the task. We use a background rejection ratio of 0.1. Threshold time between steps ($\gamma$) is chosen as 2 seconds. A step is sampled from each sub-segment based on distance to the cluster center. For visualization we display top10/top20 key-steps based on the distance to center for each sampled key step. For KS extraction results on the Ikea dataset, we show the crop around the person instead of the whole frame for clarity. The models were trained and evaluated were \emph{not} evaluated using the person-crop. 

\section{Overall flow}
\label{sec:overallflow}
\begin{algorithm}[!t]
\footnotesize
\caption{\label{alg:overallflow} Overall flow}
\begin{algorithmic}
    \STATE \textbf{Input:} Video dataset $V$
    \STATE \textbf{Output:} Adapted features $\tilde{q}$ and Key Steps $a_k$ for each video
    \STATE
    \STATE \textcolor{gray}{\# 1. Extract and store raw features}
    \STATE $P_i = \texttt{FeatureExtractor}_i(V)$
    \STATE
    \STATE \textcolor{gray}{\# 2. Train STEPs model using pre-extracted raw features}
    \STATE \textbf{for} epoch $ep = 1, \ldots, L$ \textbf{do}
        \STATE $~~~~$ \textcolor{gray}{\# Temporally sample features and create a minibatch}
        \STATE $~~~~$ $p_1, \cdots, p_i = \texttt{TemporalSampling}(P_1, \cdots, P_i)$
        \STATE $~~~~$ \textcolor{gray}{\# Forward pass through per-modality temporal encoders}
        \STATE $~~~~$ $\tilde{q}_1, \cdots, \tilde{q}_i = \texttt{f}_1(p_1), \cdots, \texttt{f}_i(p_i)$
        \STATE $~~~~$ \textcolor{gray}{\# Project features using per-modality MLP and L2 normalize}
        \STATE $~~~~$ $q_1, \cdots, q_i = \texttt{Project}_1(\tilde{q}_1) , \cdots, \texttt{Project}_i(\tilde{q}_i)$
        \STATE $~~~~$ \textcolor{gray}{\# Obtain bootstrap window using $\sigma$-window and raw features}
        \STATE $~~~~$ $\tilde{\mathcal{W}} = $ \texttt{BootstrapWindow($p_i$,$\sigma$)}
        \STATE $~~~~$ \textcolor{gray}{\# Calculate BMC2 loss and backpropagate}
        \STATE $~~~~$ \texttt{loss $=$ BMC2($q_1,\cdots,q_i,\tilde{\mathcal{W}}$)}
    \STATE \textbf{end for}
    \STATE
    \STATE \textcolor{gray}{\# 3. Evaluate for downstream task}
     \STATE \textcolor{gray}{Extract Key Steps $\{a_k\}$ for video $v$ using learned temporal encoder $f_i$}
    \STATE $ \{a_k\} = \texttt{KeySteps}(v, f_i)$ 
    \STATE \textcolor{gray}{or evaluate for Key Step Localization on dataset $V$ using learned temporal encoder $f_i$}
    \STATE \texttt{IoU}, \texttt{F1} $= \texttt{KeyStepLocalization}(V,f_i)$ 
\STATE return $\{a_k\}_{k=1}^{k=K}$, $\tilde{q}$
\end{algorithmic}
\end{algorithm}

We illustrate the overall flow of our approach in ~\cref{alg:overallflow}.

\section{Limitations and Future work}
\label{sec:limitations_and_future_work}
Our approach is the first step towards Key Step extraction for AR/VR applications where many of the modalities are available for free through on device sensors/modules. While this is generally true, extracting and parsing these modalities requires pre-trained feature extractors and/or domain knowledge. To reduce storage requirements, we work with average pooled features which precludes the model from using spatial attention. Next, while our approach can work with even a few videos, the performance gap suggests that incorporating recent advances in few-shot learning, or use of highly contextualized embeddings like CLIP can help further improve the performance. Finally, Key steps for a task can be very subjective and vary based on application. When deploying these models, key steps might have to be validated via device usage experiments and the model appropriately tuned.  

\section{Negative Societal Impact}
\label{sec:negative_impact}
The paper proposes an approach to extract key steps for unlabeled procedural videos. As such, we do not perceive any negative ethical or societal impact since experiments were done on using publicly available datasets and models. That said, while deploying such models in the wild, consent of all individuals must be taken to avoid leaking any potentially sensitive information. 
 \fi

\end{document}